\documentclass[10pt,journal,compsoc]{IEEEtran}
\usepackage[pdftex, pagebackref, colorlinks, citecolor=blue, linkcolor=blue]{hyperref}
\usepackage{eqnarray,amsmath}
\usepackage{amssymb}
\usepackage{array}
\usepackage{bm}
\usepackage{multirow}
\usepackage{graphicx} 
\usepackage{overpic}
\usepackage[dvipsnames]{xcolor}
\usepackage{subfigure}
\usepackage{url}
\usepackage{cite}
\usepackage{algorithmicx}
\usepackage[ruled]{algorithm}
\usepackage{amsthm} 
\usepackage{bm}
\usepackage{caption}
\usepackage{enumerate}
\usepackage{algpseudocode}
\usepackage{times}
\usepackage{color}
\usepackage{multirow}

\usepackage{booktabs}

\begin{document}
%
\title{Simultaneous Fidelity and Regularization Learning for Image Restoration}
%
%
%
%

\author{Dongwei~Ren,
        Wangmeng~Zuo,
        David~Zhang,
        Lei~Zhang,
        and~Ming-Hsuan Yang

\thanks{D. Ren is with the College of Intelligence and Computing, Tianjin University, Tianjin, 300350, China, and also with the Department of Computing, Hong Kong Polytechnic University, Hong Kong. E-mail: rendongweihit@gmail.com}
\thanks{W. Zuo is with the School of Computer Science and Technology, Harbin Institute of Technology, Harbin 150001, China. E-mail: cswmzuo@gmail.com}
\thanks{D. Zhang is with the School of Science and Engineering, Chinese University of Hong Kong (Shenzhen), China. Email: csdzhang@comp.polyu.edu.hk}
\thanks{L. Zhang is with the Department of Computing, Hong Kong Polytechnic
University, Hong Kong. E-mail: cslzhang@comp.polyu.edu.hk}
\thanks{M.-H. Yang is with the School of Engineering, University of California, Merced, CA 95344. E-mail: mhyang@ucmerced.edu}
\thanks{(Corresponding author: Wangmeng Zuo)}
}

\IEEEcompsoctitleabstractindextext{%
\begin{abstract}
 Most existing non-blind restoration methods are based on the assumption that a precise degradation model is known.
 As the degradation process can only be partially known or inaccurately modeled, images may not be well restored.
 Rain streak removal and image deconvolution with inaccurate blur kernels are two representative examples of such tasks.
For rain streak removal, although an input image can be decomposed into a scene layer and a rain streak layer, there exists no explicit formulation for modeling rain streaks
and the composition with scene layer.
For blind deconvolution, as estimation error of blur kernel is usually introduced, the subsequent non-blind deconvolution process does not restore the latent image well.
 %
 In this paper, we propose a principled algorithm within the maximum a posterior framework
 to tackle image restoration with a partially known or inaccurate degradation model.
 %
 Specifically, the residual caused by a partially known or inaccurate degradation model is spatially dependent and complexly distributed.
 With a training set of degraded and ground-truth image pairs, we parameterize and learn the fidelity term for a degradation model in a task-driven manner.
 %
 %
 %
 Furthermore, the regularization term can also be learned along with the fidelity term, thereby forming a simultaneous fidelity and regularization learning model.
 Extensive experimental results demonstrate the effectiveness of the proposed model for image deconvolution with inaccurate blur kernels, deconvolution with multiple degradations and rain streak removal.
\end{abstract}

}

\maketitle

\IEEEdisplaynotcompsoctitleabstractindextext
\IEEEpeerreviewmaketitle

\IEEEraisesectionheading{\section{Introduction}\label{sec:introduction}}

\IEEEPARstart{I}{mage} restoration that aims to recover
the latent clean image from a degraded observation is a fundamental problem in low-level vision.
%
%
%
However, the degradation generally is irreversible, making image restoration an ill-posed inverse problem.
While significant advances have been made in the past decades, it is challenging to develop proper models for various image restoration tasks.
%
%

In general, the linear degradation process of a clean image $\mathbf{x}$ can be modeled as
\begin{equation}
 \label{eq:general degradation model}
\mathbf{y} = \mathcal{A}\mathbf{x} + \mathbf{n},
\end{equation}
where $\mathbf{n}$ is additive noise, $\mathcal{A}$ is degradation operator, and $\mathbf{y}$ is degraded observation.
By changing the settings of the degradation operator and noise type,
they can be applied to different image restoration tasks.
%
For example, $\mathcal{A}$ can be an identity matrix for denoising,
a blur kernel convolution for deconvolution, and
a downsampling operator for super-resolution, to name a few.
%
%
The maximum a posterior (MAP) model for image restoration can then be formulated as
\begin{equation} \label{eq:guassian MAP model}
\mathbf{x}=\arg\underset{\mathbf{x}}{\min} \frac{\lambda}{2}\|\mathcal{A}\mathbf{x}-\mathbf{y}\|_2^2 + \mathcal{R}(\mathbf{x}),
\end{equation}
where $\lambda$ is a trade-off parameter, $\mathcal{R}(\mathbf{x})$ is the regularization term associated with image prior, and the fidelity term is specified by degradation $\mathcal{A}$ as well as noise $\mathbf{n}$~\cite{krishnan2009fast,zoran2011learning,schmidt2014shrinkage}.
Assuming the noise $\mathbf{n}$ is additive white Gaussian, the fidelity term can be characterized by the $\ell_2$-norm.

When the degradation operator $\mathcal{A}$ is precisely known, noise and image prior models play two key roles in the MAP-based image restoration model.
Two widely-used types of noise distributions are Gaussian and Poisson.
%
Other distributions, e.g., hyper-Laplacian \cite{wright2009robust}, Gaussian Mixture Model (GMM)~\cite{meng2013robust} and Mixture of Exponential Power (MoEP)~\cite{cao2015low}, are also introduced for modeling complex noise.
For image prior, gradient-based models, e.g., total variation \cite{chambolle2004algorithm} and hyper-Laplacian distribution \cite{krishnan2009fast},
are first studied due to simplicity and efficiency.
Subsequently, patch-based \cite{zoran2011learning} and non-local similarity \cite{dabov2006image,gu2014weighted} models are developed to characterize more complex and internal dependence among image patches.
Recently, data-driven and task-driven learning methods have also been exploited to learn regularization from training images.
The approach based on fields of experts (FoE) \cite{roth2009fields} is designed to learn the distribution of filter responses on images.
Following the FoE framework, numerous discriminative learning approaches,
e.g., cascaded shrinkage field (CSF) \cite{schmidt2014shrinkage}, trainable non-linear reaction diffusion (TNRD) \cite{chen2015learning,chen2017trainable} and universal denoising network (UNET) \cite{lefkimmiatis2018universal}, use the stage-wise learning scheme to enhance the restoration performance as well as computational efficiency.
%

%
However, the precise degradation process for most restoration tasks is not known
and thus the degradation process is modeled as
\begin{equation}
\label{eq:generalized degradation model}
\mathbf{y} = \mathcal{A}\mathbf{x} + g(\mathbf{x}; \mathcal{B}) + \mathbf{n}.
\end{equation}
In the restoration stage, only the model parameter $\mathcal{A}$ is known, while in the form $g(\mathbf{x}; \cdot)$, the noise type $\mathbf{n}$ or the parameters $\mathcal{B}$
are unknown.
Here we define this problem as \emph{image restoration with partially known or inaccurate degradation models}.

\begin{figure*}[!htbp]
	\setlength{\abovecaptionskip}{1pt}
	\setlength{\belowcaptionskip}{0pt}
\centering
\begin{tabular}{cccccc}
  &
 \multirow{7}{*}{\includegraphics[width=.155\textwidth]{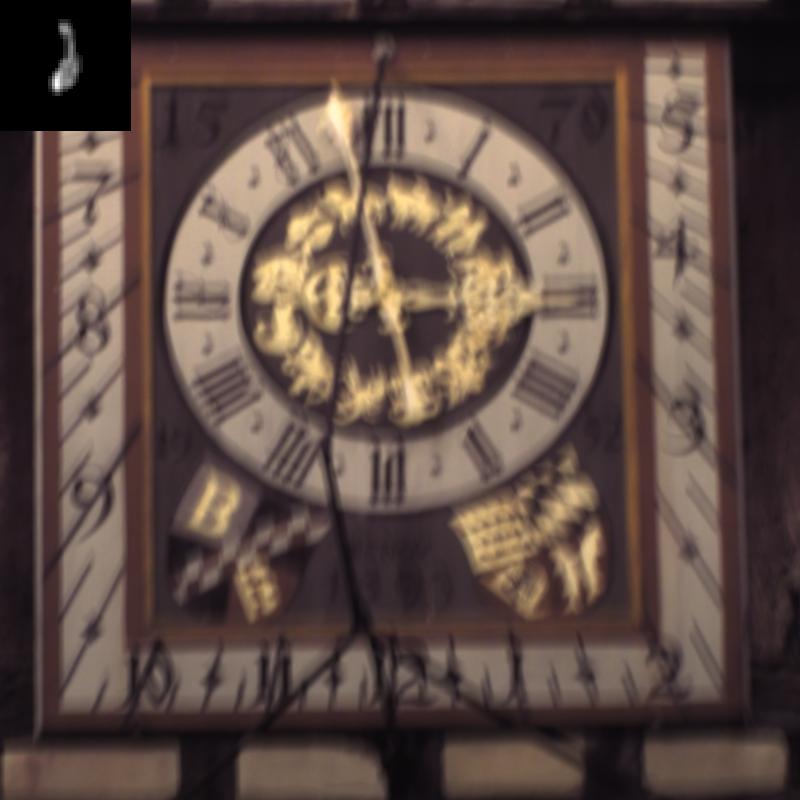}} \ &
 \multirow{7}{*}{\includegraphics[width=.24\textwidth]{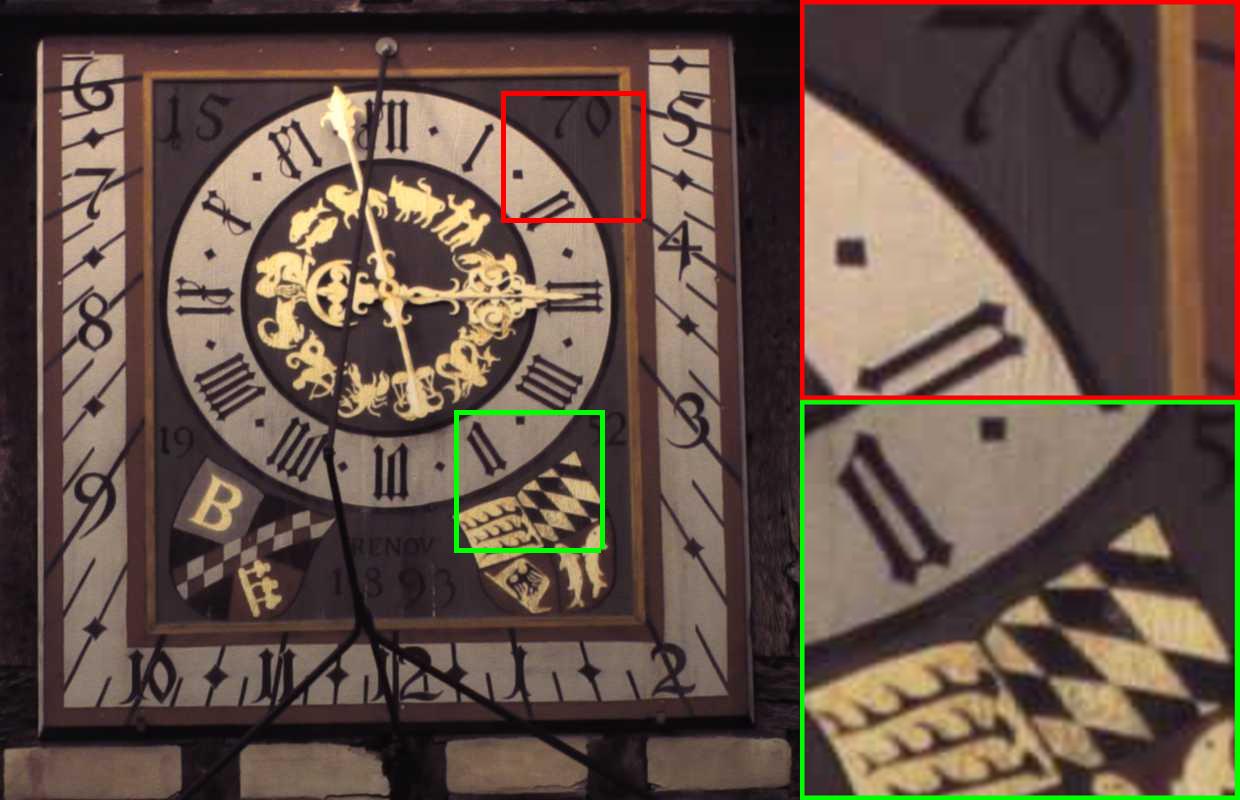}} \ &
 \multirow{7}{*}{\includegraphics[width=.24\textwidth]{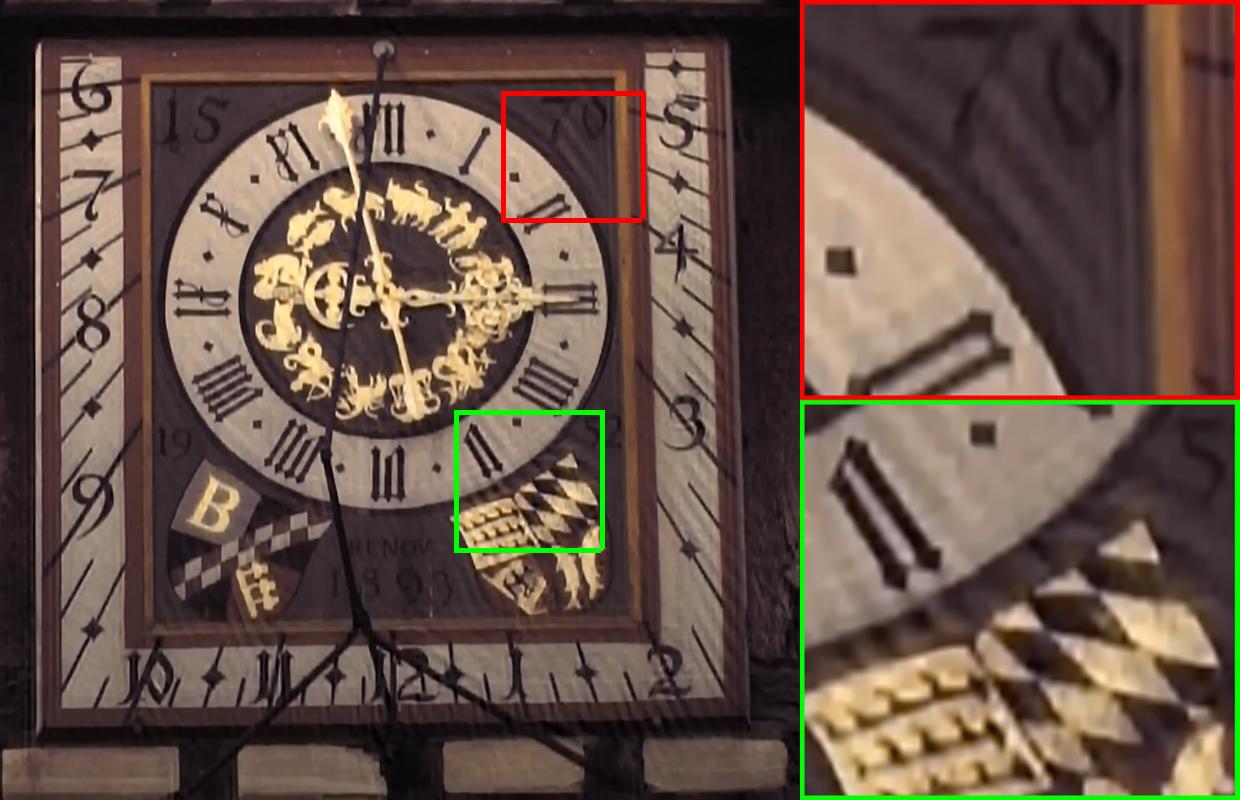}} \ &
 \multirow{7}{*}{\includegraphics[width=.24\textwidth]{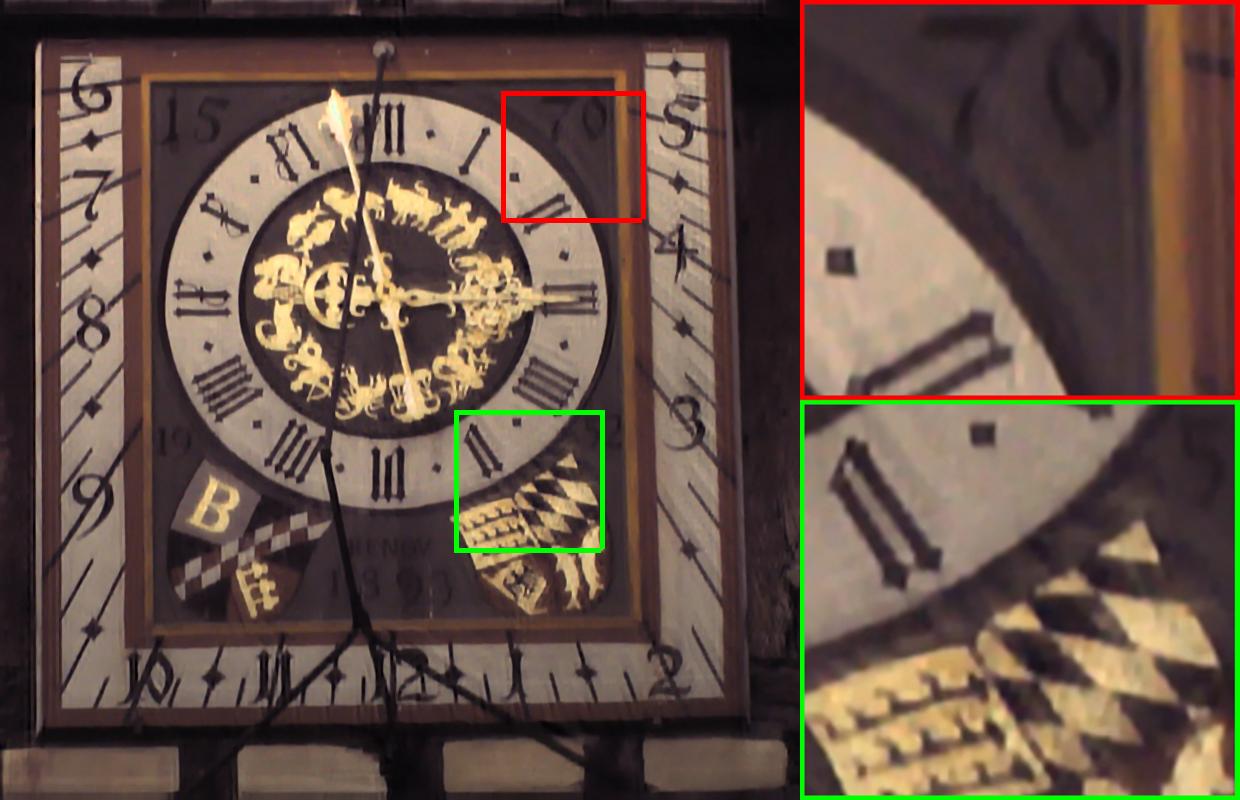}} \\
  \\
  \\
  (a)\ \ \ \\
  \\
  \\
  \\
&\footnotesize{Blurry image} & \footnotesize{Ground-truth} & \footnotesize{ROBUST \cite{ji2012robust}} & \footnotesize{SFARL} \\
  &
 \multirow{5}{*}{\includegraphics[width=.1775\textwidth]{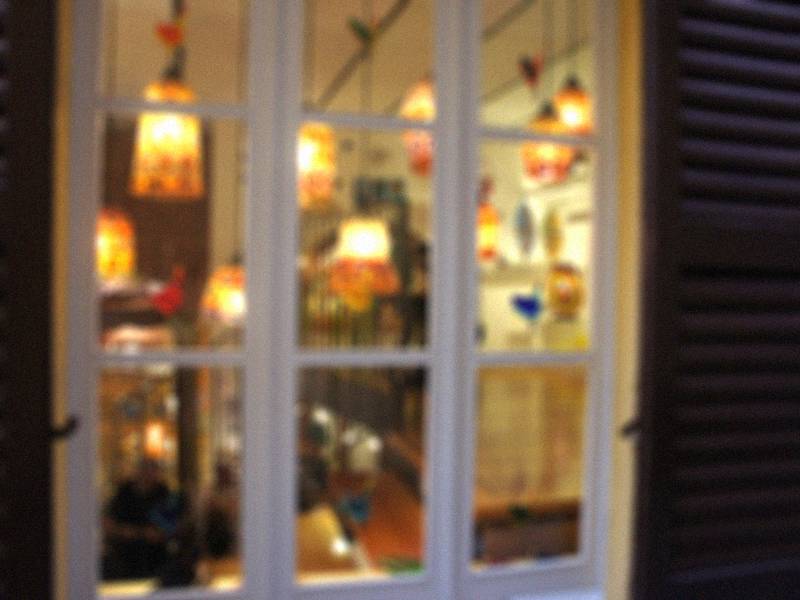}} \ &
 \multirow{5}{*}{\includegraphics[width=.24\textwidth]{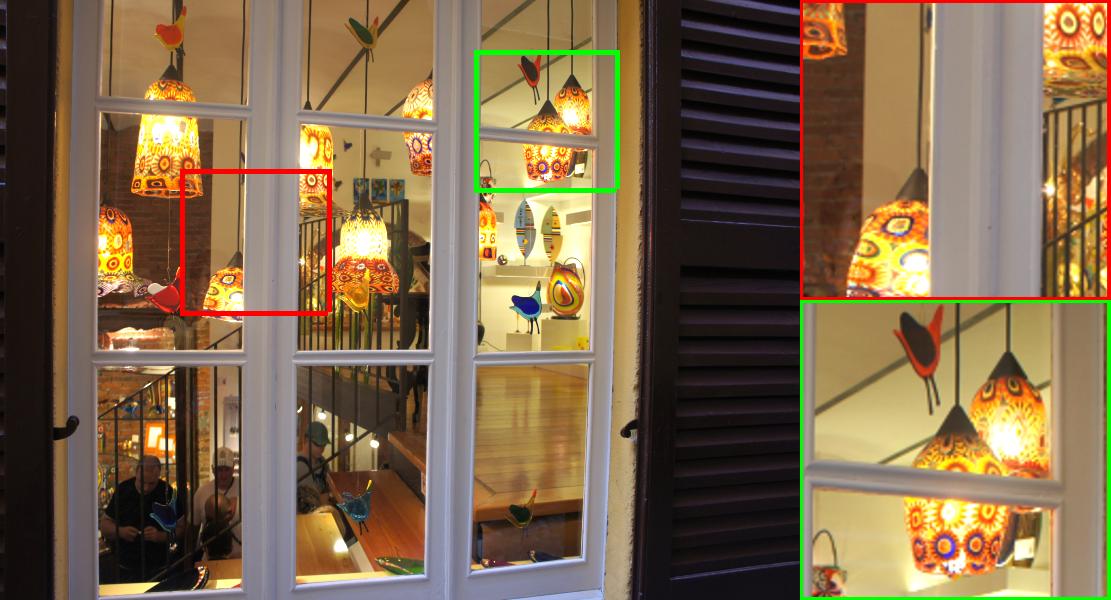}} \ &
 \multirow{5}{*}{\includegraphics[width=.24\textwidth]{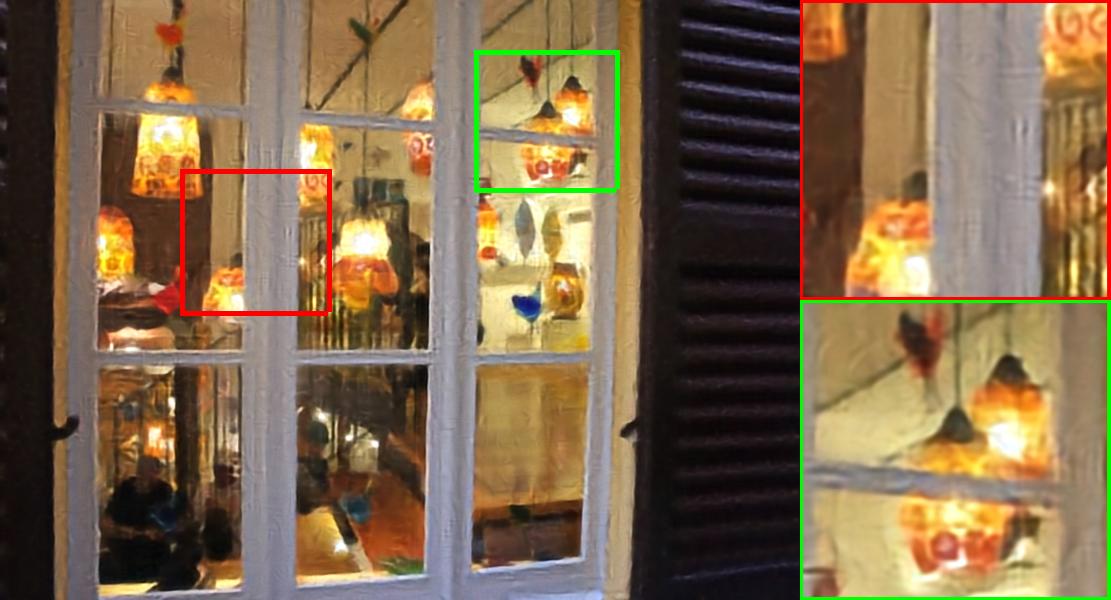}} \ &
 \multirow{5}{*}{\includegraphics[width=.24\textwidth]{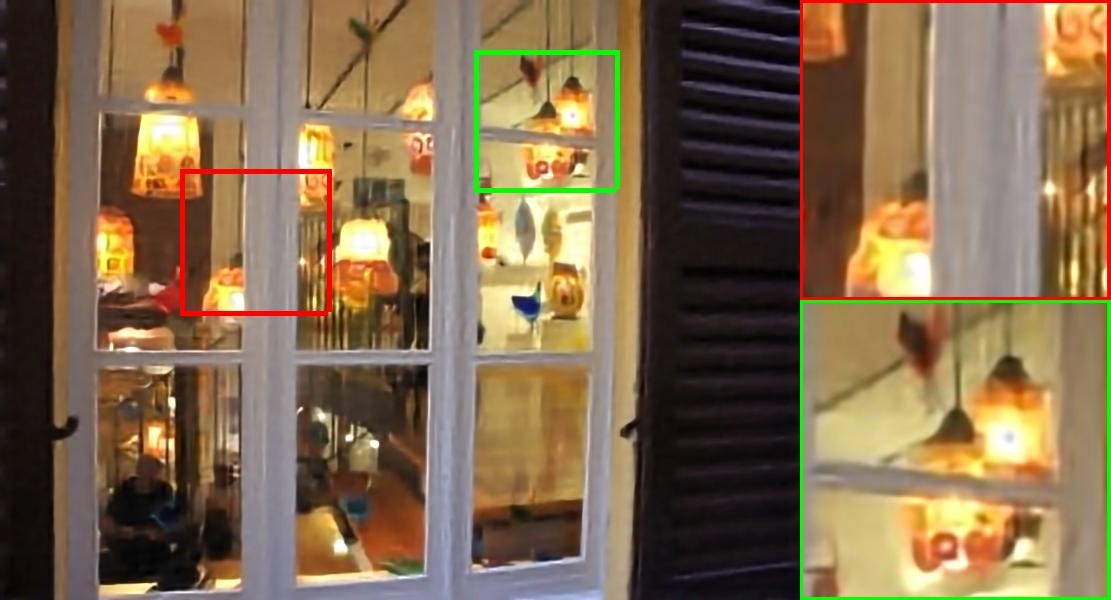}} \\
 \\
 (b)\ \ \ \\
 \\
 \vspace{10pt}\\
&\footnotesize{Degraded image} & \footnotesize{Ground-truth} & \footnotesize{DCNN\cite{xu2014deep}} & \footnotesize{SFARL}\\
  &
 \multirow{5}{*}{\includegraphics[width=.174\textwidth]{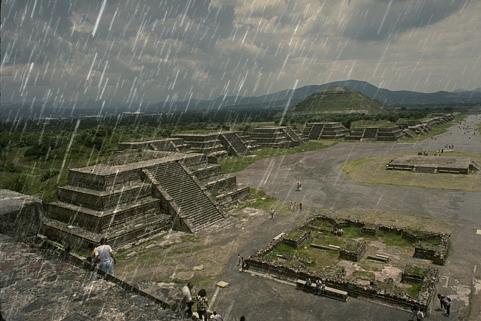}} \ &
 \multirow{5}{*}{\includegraphics[width=.24\textwidth]{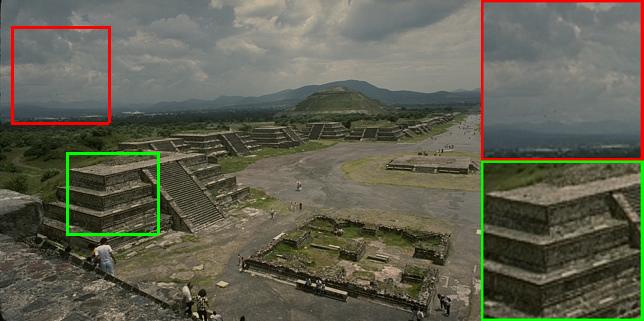}} \ &
 \multirow{5}{*}{\includegraphics[width=.24\textwidth]{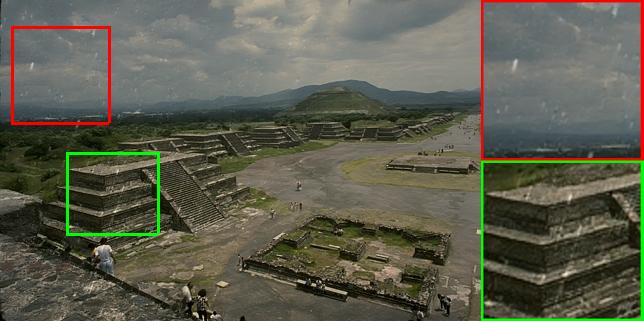}} \ &
 \multirow{5}{*}{\includegraphics[width=.24\textwidth]{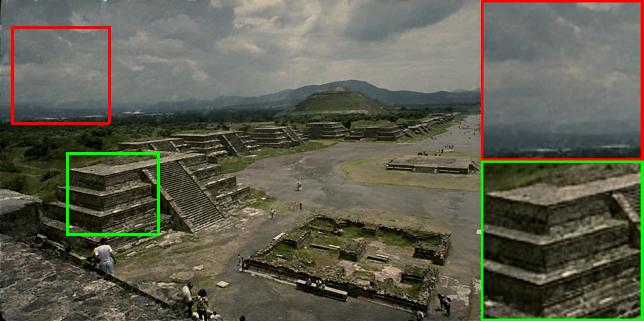}} \\
 \\
  (c)\ \ \ \\
  \\
  \vspace{3pt}\\
&\footnotesize{Rainy image} & \footnotesize{Ground-truth} & \footnotesize{DDNET\cite{fu2017removing}} & \footnotesize{SFARL} \\
\end{tabular}
\caption{\small{Illustration of the SFARL model on three restoration tasks. (a) In image deconvolution with inaccurate blur kernels, The SFARL method is effective in relieving the ringing artifacts. (b) For deconvolution along with saturation, Gaussian noise and JPEG compression, the SFARL model can achieve visually plausible result with less noises than DCNN \cite{xu2014deep}. (c) For rain streak removal, the SFARL model can produce more clean image than DDNET \cite{fu2017removing}.
}}
\label{fig:SCARL structure}
\end{figure*}

Image deconvolution with inaccurate blur kernels and rain streak removal are two representative image restoration tasks with partially known or inaccurate degradation models.
Image deconvolution with an inaccurate blur kernel is a subproblem of blind deconvolution which generally includes blur kernel estimation and non-blind deconvolution.
In the blur kernel estimation stage, the kernel error $\triangle \mathbf{k}$ generally is inevitable to be introduced by a specific method \cite{goldstein2012blur,xu2010two,zuo2016learning,ren2016image,pan2016blind,pan2017l_0}.
In the non-blind deconvolution stage, the degradation model can then be written as
\begin{equation} \label{eq:degradation_deconvolution}
\mathbf{y} = \mathbf{k} \otimes \mathbf{x} + \triangle \mathbf{k} \otimes \mathbf{x} + \mathbf{n},
\end{equation}
where $\otimes$ denotes the 2D convolution operator.
Thus, the subproblem in the non-blind deconvolution stage
is equivalent to {\emph{image deconvolution with inaccurate blur kernels}}.
Based on~\eqref{eq:generalized degradation model},
we have $g(\mathbf{x}; \triangle \mathbf{k}) = \triangle \mathbf{k} \otimes \mathbf{x}$, but $\triangle \mathbf{k}$ is unknown.
Existing non-blind deconvolution methods are sensitive to kernel error
and usually result in ringing and other artifacts \cite{krishnan2009fast,zoran2011learning}, as shown in Figure \ref{fig:SCARL structure}.

For rain streak removal, an input image $\mathbf{y}$ can be represented as the composition of a scene image layer $\mathbf{x}$ and a rain streak layer $\mathbf{x}_r$.
However, it remains challenging to model rain streak with any explicit formulation.
On one hand, a linear summation $\mathbf{y} = \mathbf{x} + \mathbf{x}_r$ is usually used for combining the scene image and rain streak layers \cite{chen2013generalized,li2016rain}.
%
%
On the other hand, it has been suggested \cite{luo2015removing} that a complex model
based on screen blend is more effective
for combining the scene image and rain streak layers,
\begin{equation} \label{eq:degradation_rain}
\mathbf{y} = \mathbf{x} - \mathbf{x} \cdot \mathbf{x}_r + \mathbf{x}_r,
\end{equation}
where $\cdot$ denotes the element-wise product.
By setting $g(\mathbf{x}; \mathbf{x}_r) = - \mathbf{x} \cdot \mathbf{x}_r$,
rain streak removal can be treated as an image restoration problem with a partially known degradation model, i.e., both $g(\mathbf{x}; \mathbf{x}_r)$ and $\mathbf{x}_r$ cannot be explicitly modeled in the deraining stage.
As shown in Figure \ref{fig:SCARL structure}, the method~\cite{li2016rain} is less effective for modeling rainy scenes, resulting in an over-smooth image with visible streaks.

Image restoration with partially known or inaccurate degradation models cannot be simply addressed by noise modeling.
From \eqref{eq:generalized degradation model}, we define the residual image as
\begin{equation}
\label{eq:residual}
\mathbf{r} = \mathbf{y} - \mathcal{A}\mathbf{x} = g(\mathbf{x}; \mathcal{B}) + \mathbf{n}.
\end{equation}
Due to the introduction of $g(\mathbf{x}; \mathcal{B})$, even $\mathbf{n}$ is white, the residual $\mathbf{r}$ is spatially dependent and complexly distributed.
Although several noise models have been suggested for complex noise modeling,
these are all based on the independent and identically distributed (i.i.d.)
assumption and ineffective for modeling the spatial dependency of the residual.
Furthermore, the characteristics of $\mathbf{r}$ is task specific and there exists no universal model that can be applied to all problems, thereby making
it more challenging to solve \eqref{eq:residual}.
%

%
%
%
%

Recently, deep CNN-based methods have achieved considerable progress on some low level vision tasks \cite{nah2017deep,su2017deep,pan2018learning,kim2016accurate,getreuer2018blade}, e.g., rain streak removal \cite{fu2017removing,he2016deep,yang2017deep}, non-blind deconvolution \cite{zhang2017learning,vasu2018non,xu2014deep} and Gaussian denosing \cite{zhang2017beyond}.
These CNN methods, however, either do not take partially known degradations into consideration, or simply address this issue by learning a direct mapping from degraded image to ground-truth.
In comparison with CNN-based models, we aim at providing a principled restoration framework for handling partially known or inaccurate degradations.


In this paper, we propose a principled fidelity learning algorithm for image restoration with partially known or inaccurate degradation models.
For either kernel error caused by a specific kernel estimation method or rain streaks, the resulting residual $\mathbf{r}$ is not entirely random
and can be characterized by spatial dependency and distribution models.
Thus, a task-driven scheme is developed to learn the fidelity term from a training set of  degraded and ground-truth image pairs.
For modeling spatial dependence and complex distribution, the residual $\mathbf{r}$ is characterized by a set of nonlinear penalty functions based on filter responses, leading to a parameterized formulation of the fidelity term.
%
Such a fidelity term is effective and flexible in modeling complex residual patterns and spatial dependency caused by partially known or inaccurate degradation for a variety of
image restoration tasks.
Furthermore, for different tasks (e.g., rain streak removal and image deconvolution), the residual patterns are also different.
With task-driven learning, the proposed method can adaptively tailor the fidelity term to specific inaccurate or partially known degradation models.

We show that the regularization term can be parameterized and learned along with the fidelity term, resulting in our simultaneous fidelity and regularization learning (SFARL) model.
In addition, we characterize the regularizer by a set of nonlinear penalty functions on filters responses of clean image.
The SFARL model is formulated as a bi-level optimization problem where
a gradient descent scheme is used to solve the inner task and stage-wise parameters are learned from the training data.
Experimental results on image deconvolution and rain streak removal
demonstrate the effectiveness of the SFARL model in terms of quantitative metrics and visual quality (see Figure \ref{fig:SCARL structure}(a)(b)(c)).
%
%
Furthermore, for image restoration with precise degradation process, e.g., non-blind Gaussian denoising, the SFARL model can be used to learn the proper fidelity term for optimizing visual perception metrics, and obtain results with better visual quality (see the results in the supplementary material).

In CSF \cite{schmidt2014shrinkage}, TNRD \cite{chen2017trainable}, and UNET \cite{lefkimmiatis2018universal}, similar parametric formulation has been adopted to model natural image prior, and discriminative learning is employed to boost restoration performance.
However, the degradation in these methods is assumed as precisely known, and thus the fidelity term is explicitly specified, e.g., $\ell_2$-norm for deconvolution with ground-truth kernel.
But in practical applications, the degradation process is usually partially
known, e.g., inaccurately estimated blur kernel, separation of rain layer and background
layer and combination of multiple degradations.
In comparison, our SFARL model aims at providing a principled restoration framework, in which fidelity term is flexible and effective to model partially known degradation
and can be jointly learned with the regularization terms during training.
As a result, when applied to image restoration with partially known or inaccurate degradation models, SFARL can be trained to perform favorably in comparison with TNRD and the state-of-the-arts.

The contributions of this work are summarized as follows:
\begin{itemize}
\item
We propose a principled algorithm for image restoration with partially known or inaccurate degradation.
Give an image restoration task, our model can adaptively learn the proper fidelity term from the training set for modeling the spatial dependency and highly complex distribution
of the task-specific residual caused by partially known or inaccurate degradation.

\item
We present a bi-level optimization model for simultaneous learning of the fidelity term as well as regularization term, and stage-wise model parameters for task-specific image restoration.
%
\item
We carry out experiments on rain streak removal, image deconvolution with inaccurate blur kernels and deconvolution with multiple degradations to
validate the effectiveness of the SFARL model.
%

\end{itemize}

\section{Related Work}
\label{sec:related}
For specific vision tasks, numerous methods have been proposed
for image deconvolution with inaccurate blur kernels and rain streak removal.
However, considerably less effort has been made to address image restoration with partially known or inaccurate degradation models.
In this section, we review related topics most relevant to this work, including noise modeling, discriminative image restoration, image deconvolution with inaccurate blur kernels, and rain streak removal.


\subsection{Noise Modeling}
For vision tasks based on robust principal component analysis (RPCA) or low rank matrix factorization (LRMF), noise is often assumed to be sparsely distributed and
can be characterized by $\ell_p$-norms \cite{wright2009robust,eriksson2010efficient}.
%
However, the noise in real scenarios is usually
more complex and cannot be simply modeled using $\ell_p$-norms.
Consequently, GMM and its variants have been used as universal approximations for modeling complex noise.
In RPCA models, Zhao et al. \cite{zhao2014robust} use a GMM model
to fit a variety of noise types, such as Gaussian, Laplacian, sparse noise and their combinations.
For LRMF, GMM is used to approximate unknown noise, and its effectiveness has been validated in face modeling and structure from motion \cite{meng2013robust}.
In addition, a GMM model is also extended for noise modeling
by low rank tensor factorization \cite{chen2016robust}, and generalized to
the Mixture of exponential power (MoEP) scheme
\cite{cao2015low} for modeling complex noise.
To determine the parameters of a GMM model,
the Dirichlet
process has been suggested to estimate the number of Gaussian components
under variational Bayesian framework \cite{chen2015bayesian}.
Recently, the weighted mixture of $\ell_1$-norm, $\ell_2$-norm \cite{gong2014image} and Gaussian \cite{zhu2016noise,xu2016patch} models have also been used for
blind denoising with unknown noise.

However, noise modeling cannot be readily used to address image restoration with partially known or inaccurate degradation models.
The residual $\mathbf{r}$ caused by inaccurate degradation is not i.i.d.
Thus, both spatial dependency and complex noise distribution need to be considered to characterize the residual.

\subsection{Discriminative Image Restoration}
%
In a MAP-based image restoration model, the regularization term is associated with a statistical prior and assumed to be learned solely based on clean images
in a generative manner, e.g.,
K-SVD \cite{aharon2006ksvd}, GMM \cite{zoran2011learning}, and FoE \cite{roth2009fields}.
Recently, discriminative learning has been extensively studied in image restoration.
In general, discriminative image restoration aims to learn a fast inference procedure by optimizing an objective function using a training set of
the degraded and ground-truth image pairs.
One typical discriminative learning approach is to combine existing image prior models with truncated optimization procedures \cite{schmidt2013discriminative,schmidt2016cascades}.
For example, CSF \cite{schmidt2014shrinkage,xiao2016learning} uses truncated half-quadratic optimization
to learn stage-wise model parameters of a modified FoE.
On the other hand, TNRD \cite{chen2015learning,chen2017trainable} unfolds a fixed number of
gradient descent inference steps.
Non-parametric methods, such as regression tree fields (RTF) \cite{schmidt2013discriminative,schmidt2016cascades} and filter forests \cite{ryan2014filter},
are also used for modeling image priors.

Existing discriminative image restoration methods, however, are all based on the precise degradation assumption.
These algorithms focus on learning regularization terms in a discriminative framework such that the models can be applied to arbitrary images and blur kernels.
In contrast, we propose a discriminative learning algorithm that considers both fidelity and regularization terms, and apply it to
image restoration with partially known or inaccurate degradation models.
%


\subsection{Image Deconvolution with Inaccurate Blur Kernels}
Typical blind deconvolution approaches consist of two stages: blur kernel estimation and non-blind deconvolution.
Existing methods mainly focus on the first stage \cite{xu2010two,zuo2016learning,cho2009fast,pan2017l_0}, and considerable attention
has been paid to blur kernel estimation.
For the second stage, conventional non-blind deconvolution methods usually are used to restore the clean image based on the estimated blur kernels.
Despite significant progress has been made in blur kernel estimation, errors are
inevitable introduced after the first stage.
Furthermore, non-blind deconvolution methods are not robust to kernel errors,
and artifacts are likely to be introduced or exacerbated
during deconvolution  \cite{krishnan2009fast,zoran2011learning}.

One intuitive solution is to design specific image priors to suppress artifacts \cite{shan2008high,perrone2012image,yuan2008progressive,heide2013high}.
%
%
%
To the best of our knowledge, there exists only one attempt \cite{ji2012robust} to implicitly model kernel error in fidelity term,
\begin{equation}\label{eq:deconvolution robust}
\mathbf{x}=\arg\underset{\mathbf{x}}{\min} \frac{\lambda}{2}\|\mathbf{k} \otimes \mathbf{x}-\mathbf{y} + \mathbf{z}\|^2 + \mathcal{R}(\mathbf{x}) + \tau\|\mathbf{z}\|_1.
\end{equation}
Here the residual $\mathbf{r}$ is defined as $\mathbf{r} = \mathbf{z} + \mathbf{n}$, where $\mathbf{z}$ is associated with the $\ell_1$-norm,
and $\mathbf{n}$ is additive white Gaussian noise.
However, a method based on $\mathbf{z}$
with the $\ell_1$-norm does not model the spatial dependency of residual signals.
The method \cite{ji2012robust} alleviates the effect of kernel errors at the expense of
potential over-smoothing restoration results.
A recent deep CNN-based approach, i.e., FCN \cite{vasu2018non}, receives multiple inputs with complementary information to produce high quality restoration result. But FCN relies on tuning parameters of non-blind deconvolution method to provide proper network inputs.
In this work, we  focus on the second stage of blind deconvolution, and
propose the SFARL model to characterize the kernel error of a specific kernel estimation method.




\subsection{Rain Streak Removal}
Rain streak and scene composition models are two important issues for removing rain drops from input images.
Based on the linear model $\mathbf{y} = \mathbf{x} + \mathbf{x}_r$, the MAP-based deraining model can be formulated as
\begin{equation} \label{eq:guassian rain model}
\begin{matrix}
\mathbf{x}=\arg\underset{\mathbf{x}}{\min} \frac{\lambda}{2}\| \mathbf{y}-\mathbf{x} - \mathbf{x}_r\|^2 + \mathcal{R}(\mathbf{x}) + \mathcal{Q}(\mathbf{x}_r)\\
\mbox{s.t.} \ \forall i,\ 0 \leq x_i \leq y_i,\ 0 \leq x_{ri} \leq y_i ,
\end{matrix}
\end{equation}
where $\mathcal{Q}(\cdot)$ denotes the regularization term of the rain streak layer, and the inequality constraints are introduced to obtain
non-negative solutions of $\mathbf{x}$ and $\mathbf{x}_r$ \cite{li2016rain}.
%
%
%
%
%

In \cite{chen2013generalized}, hand-crafted regularization is employed to impose smoothness on the image layer and low rank on the rain streak layer.
In \cite{li2016rain}, both image and rain streak layers are modeled as GMMs that are separately trained on clean patches and rain streak patches.
Based on the screen blend model, Luo et al. \cite{luo2015removing} use
the discriminative dictionary learning scheme to separate rain streaks by enforcing that
two layers need to share fewest dictionary atoms.
Recently, specifically designed CNN models \cite{fu2017removing,yang2017deep} have achieved progress in rain streak removal.
Instead of using explicit analytic models, the SFARL method is developed based on a data-driven learning approach to accommodate the complexity and diversity of
rain streak and scene composition models.
%
%
%

\section{Proposed Algorithm}
\label{sec:SCARL model}
We consider a class of image restoration problems, where the degradation model is partially known or inaccurate but a training set of degraded and ground-truth image pairs is available.
To handle these problems, we use a flexible model to parameterize the fidelity term caused by partially known or inaccurate degradation.
For a given problem, a task-driven learning approach can then be developed to
obtain a task-specific fidelity term from training data.

In this section, we first present our method for parameterizing the fidelity term to characterize the spatial dependency and complex distribution of the residual images.
In addition, the regularization term is also parameterized, resulting in our simultaneous fidelity and regularization learning model.
Finally, we propose a task-driven manner to learn the proposed model from training data.

\subsection{Fidelity Term}
The fidelity term is used to characterize the spatial dependency and highly complex distribution of the residual image $\mathbf{r} = g(\mathbf{x}; \mathcal{B}) + \mathbf{n}$.
%
On one hand, the popular explicit formulation, e.g., $\ell_2$-norm and $\ell_1$-norm, cannot model the complex distribution of residual image $\mathbf{r}$.
Due to the i.i.d. assumption, the existing noise modeling approaches, e.g., GMM \cite{zhao2014robust} and MoEP \cite{cao2015low}, also cannot be readily adopted to model spatial dependency in fidelity term.
On the other hand, the residual $\mathbf{r}$ generally is spatially dependent and complicatedly distributed.
Motivated by the success of discriminative regularization learning \cite{schmidt2014shrinkage,chen2015learning}, we also use a set of linear filters $\{\mathbf{p}_i\}_{i=1}^{N_f}$ with diverse patterns to model the spatial dependency in $g(\mathbf{x};\mathcal{B})$.
Moreover, due to the effect of $\mathbf{n}$ and its combination with $g(\mathbf{x};\mathcal{B})$, the filter responses $\{ \mathbf{r} \otimes \mathbf{p}_i \}_{i=1}^{N_f}$ remain of complex distribution.
Therefore, a set of non-linear penalty functions $\{\mathcal{D}_i\}_{i=1}^{N_f}$ is further introduced to characterize the distribution of filter responses.

To sum up, we propose a principled residual modeling in the fidelity term as follows,

\begin{equation} \label{eq:SCARL degradation model}
\begin{matrix}
\mathcal{F}(\mathbf{x}) = \lambda \sum\limits_{j=1}^{N}\sum\limits_{i=1}^{N_f} \mathcal{D}_i\left(\left(\mathbf{p}_i \otimes \left(\mathcal{A}\mathbf{x}-\mathbf{y}\right)\right)_j\!\right),
\end{matrix}
\end{equation}
where $\mathcal{A}$ is the degradation operator defined in \eqref{eq:general degradation model} and $\otimes$ is the 2D convolution operator.
%
%
In the proposed fidelity term, the parameters include $\Theta_f = \{\lambda, \mathbf{p}_i, \mathcal{D}_i\}_{i=1}^{N_f}$.
When $N_f=1$, $\mathbf{p}_1$ is delta function and $\mathcal{D}_1$ is the squared $\ell_2$-norm, the proposed model \eqref{eq:SCARL degradation model} is equivalent to the standard MAP-based model in \eqref{eq:guassian MAP model}.

Due to the introduction of linear filters $\{\mathbf{p}_i\}_{i=1}^{N_f}$ and penalty functions $\{\mathcal{D}_i\}_{i=1}^{N_f}$, the proposed fidelity term can describe the complex patterns in residual $\mathbf{r}$ caused by partially known or inaccurate degradation models.
Furthermore, our fidelity model is flexible and applicable to different tasks.
With proper training, it can be specified to certain image restoration tasks, such as rain streak removal, image deconvolution with inaccurate blur kernels.
It is worth noting that the fidelity term in \eqref{eq:SCARL degradation model} can be regarded as a special form of convolution layer in CNN.
	Nonetheless, the fidelity term \eqref{eq:SCARL degradation model} can retain better interpretability and flexibility in characterizing residual $\mathbf{r}$.
	In particular, the learned $\mathbf{p}_i$s  and $\mathcal{D}_i$s are closely related to the characteristics of redidual $\mathbf{r}$ (see an example in the supplementary material).
	Moreover, the distribution of $\mathbf{p}_i \otimes\mathbf{r}$ generally is much more complex, and cannot be simply characterized by ReLU and its variants in conventional CNN.

\subsection{Regularization Term}

To increase modeling capacity on image prior, the regularization term is further parameterized as
\begin{equation} \label{eq:SCARL regularization}
\mathcal{R}(\mathbf{x})= \sum\limits_{j=1}^{N}\sum\limits_{i=1}^{N_r} \mathcal{R}_i\left(\left(\mathbf{f}_i \otimes \mathbf{x}\right)_j\right),
\end{equation}
where $\mathbf{f}_i$ is the $i$-th linear filter, $\mathcal{R}_i$ is the corresponding non-linear penalty function, and $N_r$ is the number of linear filters and penalty functions for the regularization term.
The parameters for the regularization term include $\Theta_r = \{\mathbf{f}_i, \mathcal{R}_i\}_{i=1}^{N_r}$.
The proposed model is the generalization of the FoE \cite{roth2009fields} model by parameterizing the regularization term with both the filters and penalty functions.
Similar models have also been used in discriminative non-blind image restoration \cite{schmidt2014shrinkage,chen2015learning,lefkimmiatis2018universal}.

\subsection{SFARL Model}
Given a specific image restoration task, the parameters for the fidelity and regularization terms need to be specified.
As a large number of parameters are involved in $\mathcal{F}(\mathbf{x})$ and $\mathcal{R}(\mathbf{x})$, it is not feasible to
manually determine proper values.
In this work, we propose to learn the parameters of both fidelity and regularization terms in a task-driven manner.

Denote a training set of $S$ samples by $\{\mathbf{y}_s, \mathbf{x}_s^{gt}\}_{s=1}^S$,
where $\mathbf{y}_s$ is the $s$-th degraded image and $\mathbf{x}_s^{gt}$ is the corresponding ground-truth image.
The parameters $\Theta = \{\Theta_f, \Theta_r\}$ can be learned by solving the following bi-level optimization problem,
\begin{equation}
\label{eq:SCARL restoration model}
\begin{matrix}
\underset{\Theta}{\min}\mathcal{L}(\Theta^{}) = \sum\limits_{s=1}^S \ell\left(\mathbf{x}_s^{*},\mathbf{x}_s^{gt}\right)\\
\mbox{s.t.} \ \ \mathbf{x}_s^*=\arg\underset{\mathbf{x}\in \mathcal{X}}{\min} \
  \sum\limits_{j=1}^{N}\sum\limits_{i=1}^{N_r} \mathcal{R}_i\left(\left(\mathbf{f}_i \otimes \mathbf{x}\right)_j\right)\\
 \ \ \ \ + \lambda \sum\limits_{j=1}^{N}\sum\limits_{i=1}^{N_f} \mathcal{D}_i\left(\left(\mathbf{p}_i \otimes \left(\!\mathcal{A}\mathbf{x}\!-\!\mathbf{y}_s\right)\right)_j\right) ,\\
\end{matrix}
\end{equation}
where $\mathcal{X}$ is the feasible solution space.
For image deconvolution with an inaccurate blur kernel, the feasible solution is only constrained to be in real number space, i.e., $\mathcal{X} = \{\mathbf{x}\ |\ \mathbf{x} \in \mathbb{R}^N\}$.
For rain streak removal, additional constraints on the feasible solution space are required, i.e., $\mathcal{X} = \{\mathbf{x}\ |\ \forall i,\ 0 \leq x_i \leq y_i \}$, where $x_i$ (and $y_i$) is the $i$-th element of clean image $\mathbf{x}$ (and rainy image $\mathbf{y}$).
In principle, the trade-off parameter $\lambda$ can be absorbed into the non-linear transform $\mathcal{D}_i$ and removed from the model~(\ref{eq:SCARL restoration model}).
However, the trade-off between the fidelity and regularization terms cannot be easily made due to that the scales of $\mathcal{D}_i$ and $\mathcal{R}_i$ vary for different restoration tasks, thereby making it necessary to include $\lambda$ in~(\ref{eq:SCARL restoration model}).

The loss function $\ell(\cdot, \cdot)$ measures the dissimilarity between the output of the SFARL model and the ground-truth image.
One representative loss used in discriminative image restoration is based on the mean-squared error (MSE) \cite{chen2015learning},
\begin{equation} \label{eq:SCARL MSE}
\ell\left(\mathbf{x}, \mathbf{x}^{gt}\right) = \frac{1}{2}\|\mathbf{x} - \mathbf{x}^{gt}\|^2.
\end{equation}
For image restoration when the precise degradation process is known, the optimal fidelity term in terms of MSE becomes the negative log-likelihood.
The standard MAP model $\mathbf{x}=\arg\underset{\mathbf{x}}{\min} \frac{\lambda}{2}\|\mathcal{A} \mathbf{x}-\mathbf{y}\|^2 + \mathcal{R}(\mathbf{x})$ can then be used in the inner loop of the bi-level optimization task.
Thus, the MSE loss is only applicable to learning fidelity term for image restoration with partially known or inaccurate degradation models.

In this work, we use the visual perception metric, e.g., negative SSIM \cite{wang2004image,wang2011information},
as the loss function,
\begin{equation} \label{eq:SCARL SSIM}
\ell\left(\mathbf{x}, \mathbf{x}^{gt}\right) = - \text{SSIM}\left(\mathbf{x},  \mathbf{x}^{gt}\right).
\end{equation}
The reason of using negative SSIM is two-fold.
On one hand, it is known that SSIM is closely related to visual perception of image quality, and minimizing negative SSIM is expected to benefit the visual quality of restoration result.
On the other hand, even for image restoration with precise degradation process, the negative log-likelihood will not be the optimal fidelity term when the negative SSIM loss is used.
Thus the residual model (\ref{eq:SCARL degradation model}) can be utilized to learn proper fidelity term from training data for either image deconvolution with inaccurate blur kernels, rain streak removal, or Gaussian denoising.
In addition, the experimental results also validate the effectiveness of negative SSIM and residual modeling in terms of both visual quality and perception metric.

\section{SFARL Training}
In this section, we first present an iterative solution to inner task in the bi-level optimization problem.
The SFARL model is then parameterized and gradient-based optimization algorithm can be used for training.
The SFARL model is trained by sequentially performing greedy training in Algorithm \ref{algm:greedy training} and joint fine-tuning in Algorithm \ref{algm:joint training}.
Finally, the derivations of gradients for the greedy and end-to-end training processes are presented.

\subsection{Iterative Solution to Inner Optimization Task}
The inner task in \eqref{eq:SCARL restoration model} implicitly defines a function $\mathbf{x}^{*}(\Theta)$ on the model parameters.
As the optimization problem is non-convex, it is difficult to obtain the explicit analytic form of either $\mathbf{x}^{*}(\Theta)$ or $\frac{\partial \mathbf{x}^{*}(\Theta)} {\partial \Theta}$.
In this work, we learn $\Theta$ by considering the truncation of an iterative optimization algorithm \cite{schmidt2014shrinkage,xiao2016learning,chen2015learning,chen2017trainable}.
Furthermore, the stage-wise model parameters are also used to improve image restoration\cite{schmidt2014shrinkage,chen2015learning}.

To solve \eqref{eq:SCARL restoration model},
the updated solution $\mathbf{x}^{t+1}$ can then be written as a function of $\mathbf{x}^{t}$ and $\Theta$, i.e., $\mathbf{x}^{t+1}(\Theta; \mathbf{x}^{t})$.
Suppose that $\{(\Theta^{1}, \mathbf{x}^{1}), ..., (\Theta^{t}, \mathbf{x}^{t}) \}$ are known.
The stage-wise parameters $\Theta^{t+1}$ can then be learned by solving the following problem,
\begin{equation} \label{eq:SCARL greedy}
\underset{\Theta^{t+1}}{\min}\mathcal{L}(\Theta^{t+1}) = \sum\limits_{s=1}^S \ell\left(\mathbf{x}_s^{t+1}(\Theta^{t+1}; \mathbf{x}^{t}),\mathbf{x}_s^{gt}\right).
\end{equation}
Here we use a gradient descent method to solve the inner optimization loop,
and $\mathbf{x}^{t+1}(\Theta; \mathbf{x}^{t})$ can be written as
%
%
\begin{equation} \label{eq:SCARL gradient model}
\begin{aligned}
\mathbf{x}^{t+1}(\Theta^{t+1}; \mathbf{x}^{t})&= \mathbf{x}^{t} - \sum\limits_{i=1}^{N_r} \bar{\mathbf{f}}_i^{t+1}\!\otimes \!\phi^{t+1}_i(\mathbf{f}^{t+1}_i \! \otimes \! \mathbf{x}^{t})\\
&\!\!\!\!\!\!\!\!\!\!\!\!\!\!\!\!\!\!\!\!\!\!\!\! - \lambda^{t+1}\! \sum\limits_{i=1}^{N_f} \mathcal{A}^{\top} \bar{\mathbf{p}}_i^{t+1}\!\!\otimes\varphi^{t+1}_i\!\left({\mathbf{p}}^{t+1}_i \! \otimes \! \left(\mathcal{A}\mathbf{x}^{t}- \mathbf{y}\right)\right),
\end{aligned}
\end{equation}
where the influence functions are defined as $\varphi_i = \mathcal{D'}_i$ and $\phi_i = \mathcal{R'}_i$.
These functions are entry-wisely performed on a vector or matrix.
In addition, $\bar{\mathbf{p}}_i$ and $\bar{\mathbf{f}}_i$ are filters by rotating ${\mathbf{p}_i}$ and ${\mathbf{f}_i}$ 180 degrees, respectively.
After each gradient descent step, $\mathbf{x}^{t+1}$ is
projected to the feasible solution space $\mathcal{X}$.
The inference procedure is shown in Algorithm \ref{algm:sfarl inference}.
%

We use ADAM \cite{kingma2014adam} to solve the optimization problem in \eqref{eq:SCARL greedy}.
Therefore, we need to present the parameterization of the solution in \eqref{eq:SCARL gradient model} and
derive the gradients for the greedy and end-to-end learning processes.

\begin{algorithm}[!ht] \footnotesize
\caption{SFARL}
\label{algm:sfarl inference}
\begin{algorithmic}[1]
 \Require Current result $\mathbf{x}^0$, degraded image $\mathbf{y}$, degradation operator $\mathcal{A}$, model parameters $\{\Theta^{t}\}_{t=1}^{T}$
 \Ensure Restoration results $\mathbf{x}$
\For{$t = 0 \text{ to } T-1$}

 \State Compute $\mathbf{x}^{t+1}$ using \eqref{eq:SCARL gradient model}

\EndFor
\State   $\mathbf{x} = \{\mathbf{x}^{1},\mathbf{x}^{2},..., \mathbf{x}^{T}\}$
\end{algorithmic}
\end{algorithm}
\begin{algorithm}[!ht] \footnotesize
\caption{Greedy Training}
\label{algm:greedy training}
\begin{algorithmic}[1]
 \Require Training data $\{\mathbf{y}_i,\mathbf{x}_i^{gt},\mathcal{A}_i\}_{i=1}^{N}$
 \Ensure SFARL parameters $\{\Theta^{t}\}_{t=1}^{T}$
\State Set stage number $T$, epoch number $E$, mini-batch size $n$, mini-batch number $M = N/n$
\State Initialize: $\{\mathbf{x}_i^0 | \mathbf{x}_i^0 = \mathbf{y}_i\}_{i=1}^N$, $\{\Theta^{t}\}_{t=1}^{T}$

\For{$t = 0 \text{ to } T-1$}

\For{$epoch = 1 \text{ to } {E}$}

\For{$m = 0 \text{ to } M-1$}
  \State Prepare $m$-th mini-batch data: $\{\mathbf{y}_{i},\mathbf{x}_{i}^{gt},\mathcal{A}_{i}\}_{i = m\times n +1}^{m\times n + n}$
  \State Forward samples in $m$-th mini-batch: $$\mathbf{x}_i^{t+1} = \text{SFARL}(\mathbf{x}_i^{t},\mathbf{y}_i,\mathcal{A}_i,\Theta^{t+1})$$
  \State Compute gradients for stage $t+1$: $\frac{1}{n} \sum_{i} \frac{\partial \ell(\mathbf{x}_i^{t+1},\mathbf{x}_i^{gt})}{\partial \Theta^{t+1}}$
  \State Use Adam to optimize stage $t+1$ parameters $\Theta^{t+1}$
\EndFor
\EndFor
\EndFor
\end{algorithmic}
\end{algorithm}
\begin{algorithm}[!ht] \footnotesize
\caption{Joint Fine-tuning}
\label{algm:joint training}
\begin{algorithmic}[1]
 \Require Training data $\{\mathbf{y}_i,\mathbf{x}_i^{gt},\mathcal{A}_i\}_{i=1}^{N}$, model parameters $\{\Theta^{t}\}_{t=1}^{T}$
 \Ensure SFARL parameters $\{\Theta^{t}\}_{t=1}^{T}$
\State Set epoch number $E$, mini-batch size $n$, mini-batch number $M = N/n$
\State Initialize $\{\mathbf{x}_i^0 | \mathbf{x}_i^0 = \mathbf{y}_i\}_{i=1}^N$

\For{$epoch = 1 \text{ to } {E}$}

\For{$m = 0 \text{ to } M-1$}
  \State Prepare $m$-th mini-batch data: $\{\mathbf{y}_{i},\mathbf{x}_{i}^{gt},\mathcal{A}_{i}\}_{i = m\times n +1}^{m\times n +n}$
  \State Forward samples in $m$-th mini-batch:
  $$\{\mathbf{x}_i^1,\mathbf{x}_i^2,...,\mathbf{x}_i^T\} = \text{SFARL}(\mathbf{x}_i^{0},\mathbf{y}_i,\mathcal{A}_i,\{\Theta^{t}\}_{t=1}^{T})$$
  \State Compute gradients for each stage: $\{\frac{1}{n}\sum_i \frac{\partial \ell(\mathbf{x}_i^{T},\mathbf{x}_i^{gt})}{\partial \Theta^{t}}\}_{t=1}^{T}$
  \State Use Adam to end-to-end optimize parameters $\{\Theta^{t}\}_{t=1}^{T}$
  \EndFor
\EndFor

\end{algorithmic}
\end{algorithm}

\subsection{Parameterization}
Similar to \cite{schmidt2014shrinkage,chen2015learning}, we use the weighted summation of Gaussian RBF functions to parameterize the influence functions in regularization term
\begin{equation}
\phi_i(z) = \sum\limits_{j=1}^{M}{{{\pi }_{ij}}\exp \left( -\frac{\gamma }{2}{{\left( z - {{\mu }_{j}} \right)}^{2}} \right)},
\end{equation}
and in fidelity term
\begin{equation}
\varphi_i(z) = \sum\limits_{j=1}^{M}{{{w }_{ij}}\exp \left( -\frac{\gamma }{2}{{\left( z - {{\mu }_{j}} \right)}^{2}} \right)},
\end{equation}
where $\pi _{ij}$ and $w_{ij}$ are weight coefficients, $\mu_j$ is mean value and $\gamma$ is precision.

The filters $\mathbf{f}_i$ in regularization term and $\mathbf{p}_i$ in fidelity term are specified as linear combination of DCT basis with unit norm constraint,
\begin{equation}
\mathbf{f}_i = \mathcal{B}_r\frac{\mathbf{s}_i}{\|\mathbf{s}_i\|_2} \ \ \text{and}\ \ \mathbf{p}_i = \mathcal{B}\frac{\mathbf{c}_i}{\|\mathbf{c}_i\|_2},
\end{equation}
where $\mathcal{B}$ is complete DCT basis, $\mathcal{B}_r$ is DCT basis by excluding the DC component, $\mathbf{s}_i$ and $\mathbf{c}_i$ are coefficients for regularization term and fidelity term respectively.

In our implementation, we utilize filters with size $7 \times 7$ in both regularization term and fidelity term.
Thus, the numbers of non-linear functions and filters can be accordingly set, i.e., $N_r=48$ for regularization term, and $N_f=49$ for fidelity term.
The numbers of Gaussian functions are fixed to $63$ for both fidelity and regularization terms, i.e., $M=63$.
To handle the boundary condition in convolution operation, the image is padded for processing and only the valid region is cropped for output.

\subsection{Greedy Training}
The SFARL model is firstly trained stage-by-stage.
To learn the model parameters of stage $t+1$, we need to compute gradient by the chain rule,
\begin{equation} \label{eq:supp_SCARL greedy gradient}
\frac{\partial \ell(\mathbf{x}^{t+1},\mathbf{x}^{gt})}{\partial \Theta^{t+1}} =
\frac{\partial \mathbf{x}^{t+1}}{\partial \Theta^{t+1}}
\frac{\partial \ell(\mathbf{x}^{t+1},\mathbf{x}^{gt})}{\partial \mathbf{x}^{t+1}}.
\end{equation}

\subsubsection{Deviation of $\frac{\partial \ell(\mathbf{x}^{t+1},\mathbf{x}^{gt})}{\partial \mathbf{x}^{t+1}}$}
When the loss function is specified as MSE, i.e., $\ell\left(\mathbf{x}^{t+1}, \mathbf{x}^{gt}\right) = \frac{1}{2}\|\mathbf{x}^{t+1} - \mathbf{x}^{gt}\|^2$, the gradient can be simply computed as
\begin{equation} \label{eq:supp_gradient mse}
\frac{\partial \ell(\mathbf{x}^{t+1},\mathbf{x}^{gt})}{\partial \mathbf{x}^{t+1}} = \mathbf{x}^{t+1} - \mathbf{x}^{gt}.
\end{equation}

\subsubsection*{Visual perception metric, i.e., negative SSIM}
When the loss function is specified as visual perception metric, i.e., $\ell\left(\mathbf{x}^{t+1}, \mathbf{x}^{gt}\right) = -\text{SSIM}(\mathbf{x}^{t+1}, \mathbf{x}^{gt})$ \cite{wang2004image,wang2011information}, we give the gradient deviation as follows.
To distinct the entire image and small patch, only in this subsection we use $\mathbf{X}$ and $\mathbf{Y}$ as entire image and reference image respectively.
The SSIM value is computed based on the small patches $\mathbf{x}_i$ and $\mathbf{y}_i$
\begin{equation}
\text{SSIM} \left( \mathbf{X},\mathbf{Y} \right)=\frac{1}{{{N}_{s}}}\sum\limits_{i=1}^{{{N}_{s}}}{\text{ssim}\left( {{\mathbf{x}}_{i}},{{\mathbf{y}}_{i}} \right)},
\end{equation}
where $N_s$ is the number of patches.
The value on each patch is computed as
\begin{equation}
 \text{ssim}\left( \mathbf{x},\mathbf{y} \right)=\frac{\left( 2{{\mu }_{x}}{{\mu }_{y}}+{{C}_{1}} \right)\left( 2{{\sigma }_{xy}}+{{C}_{2}} \right)}{\left( \mu _{x}^{2}+\mu _{y}^{2}+{{C}_{1}} \right)\left( \sigma _{x}^{2}+\sigma _{y}^{2}+{{C}_{2}} \right)},
\end{equation}
where
${{\mu }_{x}}=\frac{1}{{{N}_{p}}}\left( {{\mathbf{1}}^{\top}}\mathbf{x} \right)$ is mean value of patch $\mathbf{x}$, $\sigma _{x}^{2}=\frac{1}{{{N}_{p}}-1}{{\left( \mathbf{x}-{{\mu }_{x}} \right)}^{\top}}\left( \mathbf{x}-{{\mu }_{x}} \right)$ is variance of patch $\mathbf{x}$, and ${{\sigma }_{xy}}=\frac{1}{{{N}_{p}}-1}{{\left( \mathbf{x}-{{\mu }_{x}} \right)}^{\top}}\left( \mathbf{y}-{{\mu }_{y}} \right)$ is covariance of pathes $\mathbf{x}$ and $\mathbf{y}$, and $C_1$, $C_2$ are some constant values.
Let us define ${{A}_{1}} = 2{{\mu }_{x}}{{\mu }_{y}}+{{C}_{1}}$, ${{A}_{2}} = 2{{\sigma }_{xy}}+{{C}_{2}}$, $B_1 = \mu _{x}^{2}+\mu _{y}^{2}+{{C}_{1}}$ and $B_2 = \sigma _{x}^{2}+\sigma _{y}^{2}+{{C}_{2}}$. Then we have $S \left( \mathbf{x},\mathbf{y} \right) = \frac{A_1 A_2}{B_1 B_2}$.

The gradient of negative SSIM is
\begin{equation}
\begin{aligned}
   \frac{\partial \left(-\text{SSIM}\left( \mathbf{X},\mathbf{Y} \right)\right)}{\partial \mathbf{X}}&=-\frac{1}{{{N}_{s}}}\sum\limits_{i=1}^{{{N}_{s}}}{\frac{\partial \left(\text{ssim}\left( {{\mathbf{x}}_{i}},{{\mathbf{y}}_{i}} \right)\right)}{\partial \mathbf{Y}}} \\
  &\!\!\!\!\!\!\!\!\!\!\!\!=-\frac{1}{{{N}_{s}}}\sum\limits_{i=1}^{{{N}_{s}}}{{{\left. \frac{\partial \left(\text{ssim}\left( \mathbf{x},\mathbf{y} \right)\right)}{\partial \mathbf{X}} \right|}_{\mathbf{x}={{\mathbf{x}}_{i}},\mathbf{y}={{\mathbf{y}}_{i}}}}} ,\\
\end{aligned}
\end{equation}
where
\begin{equation}
\begin{aligned}
\frac{\partial \left(\text{ssim}\left( \mathbf{x},\mathbf{y} \right)\right)}{\partial \mathbf{x}}&=\frac{2}{{{N}_{p}}B_{1}^{2}B_{2}^{2}}\left( {{A}_{1}}{{B}_{1}}\left( {{B}_{2}}\mathbf{x}-{{A}_{2}}\mathbf{y} \right)\right.\\
&\!\!\!\!\!\!\!\!\!\!\!\!\!\!\!\!\!\! +\left. {{B}_{1}}{{B}_{2}}\left( {{A}_{2}}-{{A}_{1}} \right){{\mu }_{x}}\mathbf{1}+{{A}_{1}}{{A}_{2}}({{B}_{1}}-{{B}_{2}}){{\mu }_{y}}\mathbf{1} \right).
\end{aligned}
\end{equation}
For simplicity, we hereafter use $\mathbf{e}$ to denote $\frac{\partial \ell(\mathbf{x}^{t+1},\mathbf{x}^{gt})}{\partial \mathbf{x}^{t+1}}$ for both MSE and negative SSIM.

\subsubsection{Deviation of $\frac{\partial \mathbf{x}^{t+1}}{\partial \Theta^{t+1}}$}
Since the parameterization of fidelity term and regularization term is similar, we only use the fidelity term as an example, and it is easy to extend it to the regularization term.

\subsubsection*{Weight parameter $\lambda$}
The gradient with respect to  $\lambda$ is
\begin{equation}\footnotesize
\frac{\partial \mathbf{x}^{t+1}}{\partial \lambda^{t+1}} = -\left(\sum\limits_{i=1}^{N_f} \mathcal{A}^{\top} \bar{\mathbf{p}}_i^{t+1}\!\!\otimes\varphi^{t+1}_i\left({\mathbf{p}}^{t+1}_i\otimes\left(\mathcal{A}\mathbf{x}^{t}-\mathbf{y}\right)\right)\right)^{\top}.
\end{equation}
The overall gradient with respect to  $\lambda$ is
\begin{equation} \footnotesize
\frac{\partial \ell(\mathbf{x}^{t+1},\mathbf{x}^{gt})}{\partial \lambda^{t+1}} = -\left(\sum\limits_{i=1}^{N_c} \mathcal{A}^{\top} \bar{\mathbf{p}}_i^{t+1}\!\!\otimes\varphi^{t+1}_i\left({\mathbf{p}}^{t+1}_i\otimes\left(\mathcal{A}\mathbf{x}^{t}
-\mathbf{y}\right)\right)\right)^{\top}\mathbf{e}.
\end{equation}

\subsubsection*{Filter $\mathbf{p}_i$}
The function $\mathbf{x}^{t+1}$ with respect to  each filter $\mathbf{p}_i$ can be simplified to,
\begin{equation}
\mathbf{x}^{t+1} = -\lambda^{t+1} \mathcal{A}^{\top} \bar{\mathbf{p}}_i^{t+1}\otimes\varphi^{t+1}_i\left({\mathbf{p}}^{t+1}_i\otimes\left(\mathcal{A}\mathbf{x}^{t}-\mathbf{y}\right)\right) + C,
\end{equation}
where $C$ denotes a constant which is independent with $\mathbf{p}_i$.
Let us define $\mathbf{u} = -\mathcal{A}^{\top} \bar{\mathbf{p}}_i^{t+1}$ and $\mathbf{v} = \varphi^{t+1}_i\left({\mathbf{p}}^{t+1}_i\otimes\left(\mathcal{A}\mathbf{x}^{t}-\mathbf{y}\right)\right)$.
Thus, we can obtain the gradient deviation as
\begin{equation} \label{eq:supp_grad x p}
\frac{\partial \mathbf{x}^{t+1}}{\partial \mathbf{p}_i^{t+1}} = \frac{\partial \mathbf{u}}{\partial \mathbf{p}_i^{t+1}} \frac{\partial \mathbf{x}^{t+1}}{\partial \mathbf{u}}
+
\frac{\partial \mathbf{v}}{\partial \mathbf{p}_i^{t+1}} \frac{\partial \mathbf{x}^{t+1}}{\partial \mathbf{v}}.
\end{equation}
Based on the convolution theorem \cite{bracewell1986fourier}, we have
\begin{equation}
\mathbf{u} \otimes \mathbf{v} \Leftrightarrow \mathbf{U}\mathbf{v} \Leftrightarrow \mathbf{V}\mathbf{u},
\end{equation}
where $\mathbf{U}$ and $\mathbf{V}$ are sparse convolution matrices of $\mathbf{u}$ and $\mathbf{v}$, respectively.
Thus, the first term in \eqref{eq:supp_grad x p} is
\begin{equation}
\frac{\partial \mathbf{u}}{\partial \mathbf{p}_i^{(t)}} \frac{\partial \mathbf{x}^{(t)}}{\partial \mathbf{u}}
=-R_{180}^{\top}\mathcal{A}\mathbf{V}^{\top},
\end{equation}
where $R_{180}$ rotates matrix by 180 degrees.

For the second term, we introduce an auxiliary variable $\mathbf{b}=\mathcal{A}\mathbf{x}^{t}-\mathbf{y}$,
$\mathbf{z} = {\mathbf{p}}^{t+1}_i\otimes\mathbf{b}$,
and we have $\mathbf{v} = \varphi_i^{t+1}\left(\mathbf{z}\right)$.
We note that
$$\mathbf{z} = {\mathbf{p}}^{t+1}_i\otimes \mathbf{b} \Leftrightarrow \mathbf{B}\mathbf{p}^{t+1}_i.$$
Therefore, we have
\begin{equation}
\frac{\partial \mathbf{v}}{\partial \mathbf{p}_i^{t+1}}
\frac{\partial \mathbf{x}^{t+1}}{\partial \mathbf{v}}
=
\frac{\partial \mathbf{z}}{\partial \mathbf{p}_i^{t+1}}
\frac{\partial \mathbf{v}}{\partial \mathbf{z}}
\frac{\partial \mathbf{x}^{t+1}}{\partial \mathbf{v}}
=
-\mathbf{B}^{\top}\Lambda\mathbf{U}^{\top},
\end{equation}
where $\Lambda =\text{diag} \left(\varphi^{t+1'}_i(z_1),...,\varphi^{t+1'}_i(z_N)\right)$ is a diagonal matrix.
The gradient of $\varphi^{}_i(z)$ is
\begin{equation}
\varphi^{'}_i(z) = -\gamma\sum\limits_{j=1}^{M}{{{w }_{ij}}\exp \left( -\frac{\gamma }{2}{{\left( z - {{\mu }_{j}} \right)}^{2}} \right)} \left(z-\mu_j\right).
\end{equation}

Since the filter is specified as linear combination of DCT basis, one need to
derive the gradient with respect to  the combination coefficients $\mathbf{c}$, i.e.,
\begin{equation}
\frac{\partial \ell}{\partial \mathbf{c}} = \frac{\partial \mathbf{p}}{\partial \mathbf{c}}
\frac{\partial \ell}{\partial \mathbf{p}}.
\end{equation}
By introducing $\mathbf{v}=\frac{\mathbf{c}}{\|\mathbf{c}\|_2}$, we then have
\begin{equation}
\begin{aligned}
\frac{\partial \mathbf{p}}{\partial \mathbf{c}} &=
\frac{\partial \mathbf{v}}{\partial \mathbf{c}}
\frac{\partial \mathbf{p}}{\partial \mathbf{v}}
=\frac{\partial \mathbf{v}}{\partial \mathbf{c}}\mathcal{B}^{\top}\\
&=\left(\frac{\mathbf{I}}{\|\mathbf{c}\|_2}
+ \frac{\partial (\mathbf{c}^{\top}\mathbf{c})^{-\frac{1}{2}}}{\partial \mathbf{c}} \right)\mathcal{B}^{\top}\\
&=\left(\frac{\mathbf{I}}{\|\mathbf{c}\|_2}
+ \frac{\partial (\mathbf{c}^{\top}\mathbf{c})}{\partial \mathbf{c}}
(-\frac{1}{2}\frac{1}{\|\mathbf{c}\|_2^3})\mathbf{c}^{\top}
\right)\mathcal{B}^{\top}\\
&=\left(\frac{\mathbf{I}}{\|\mathbf{c}\|_2}
+ 2\mathbf{c}
(-\frac{1}{2}\frac{1}{\|\mathbf{c}\|_2^3})\mathbf{c}^{\top}
\right)\mathcal{B}^{\top}\\
&=\frac{1}{\|\mathbf{c}\|_2}\left(\mathbf{I}
- \frac{\mathbf{c}}{\|\mathbf{c}\|_2}
\frac{\mathbf{c}^{\top}}{\|\mathbf{c}\|_2}
\right)\mathcal{B}^{\top}.
\end{aligned}
\end{equation}

Finally, the overall gradient with respect to  combination coefficients $\mathbf{c}_i^{t+1}$ is given by
\begin{equation} \footnotesize
\frac{\partial \ell}{\partial \mathbf{c}_i^{t+1}}\! =\!
-\frac{1}{\|\mathbf{c}_i^{t+1}\|_2}\!\!\left(\mathbf{I}
\!- \!\frac{\mathbf{c}_i^{t+1}}{\|\mathbf{c}_i^{t+1}\|_2}
\frac{(\mathbf{c}_i^{t+1})^{\top}}{\|\mathbf{c}_i^{t+1}\|_2}
\!\right)\!\mathcal{B}^{\top}\!
\left(\mathbf{B}^{\top}\Lambda\mathbf{U}^{\top}
\!+\!R_{180}^{\top}\mathcal{A}\mathbf{V}^{\top}\!\right)\!\mathbf{e}.
\end{equation}

\subsubsection*{Non-linear function $\varphi_i$}
We first reformulate the function $\mathbf{x}^{t+1}$ with respect to  $\varphi_i$ into the matrix form
\begin{equation}
\mathbf{x}^{t+1} \sim -\lambda^{t+1} \mathcal{A}^{\top} ({\mathbf{P}}_i^{t+1})^\top \varphi^{t+1}_i\left(\mathbf{b}\right),
\end{equation}
where $\mathbf{b}=\mathbf{P}^{t+1}_i\left(\mathcal{A}\mathbf{x}^{t}-\mathbf{y}\right)$.
Therefore, the column vector $\varphi^{t+1}_i(\mathbf{b})$ can be reformulated into the matrix form,
\begin{equation}
\varphi^{t+1}_i(\mathbf{b}) = \mathbf{G}(\mathbf{b})\mathbf{w}_i^{t+1},
\end{equation}
where $\mathbf{w}_i$ is the vectorized version of parameters $w_{ij}$, matrix $\mathbf{G}(\mathbf{b})$ is
\begin{equation*}\footnotesize
\!\!\!\!\mathbf{G}(\mathbf{b})\!\!=\!\!\!\left[
\begin{matrix}
\exp(-\frac{\gamma}{2}(b_1\!-\!\mu_1)^2)\!\!\!&\!\!\!\exp(-\frac{\gamma}{2}(b_1\!-\!\mu_2)^2)\!\!\!\!\!&\!\!\!\!\!\cdots\!\!\!\!\!&\!\!\!\!\!\exp(-\frac{\gamma}{2}(b_1\!-\!\mu_M)^2)\\
\exp(-\frac{\gamma}{2}(b_2\!-\!\mu_1)^2)\!\!\!&\!\!\!\exp(-\frac{\gamma}{2}(b_2\!-\!\mu_2)^2)\!\!\!\!\!&\!\!\!\!\!\cdots\!\!\!\!\!&\!\!\!\!\!\exp(-\frac{\gamma}{2}(b_2\!-\!\mu_M)^2)\\
\vdots&\vdots&\ddots&\vdots\\
\exp(-\frac{\gamma}{2}(b_N\!-\!\mu_1)^2)\!\!\!&\!\!\!\exp(-\frac{\gamma}{2}(b_N\!-\!\mu_2)^2)\!\!\!\!\!&\!\!\!\!\!\cdots\!\!\!\!\!&\!\!\!\!\!\exp(-\frac{\gamma}{2}(b_N\!-\!\mu_M)^2)\\
\end{matrix}
\right].
\end{equation*}
Thus, we can get
\begin{equation}
\frac{\partial \mathbf{x}^{t+1}}{\partial \mathbf{w}_i^{t+1}}=-\lambda^{t+1}\mathbf{G}^{\top}\mathbf{P}_i^{t+1}\mathcal{A},
\end{equation}
and finally the overall gradient with respect to  $\mathbf{w}_i^{t+1}$ is
\begin{equation}
\frac{\partial \ell}{\partial \mathbf{w}_i^{t+1}}=-\lambda^{t+1}\mathbf{G}^{\top}\mathbf{P}_i^{t+1}\mathcal{A} \mathbf{e}.
\end{equation}
In our implementation, we do not explicitly compute the matrix ${\bf U, V, B}$, since they can be efficiently operated via 2D convolution.

\subsection{Joint Fine-tuning}
Once the greedy training process for each stage is carried out,
an end-to-end training process is used to fine-tune all the parameters across stages.
The joint training loss function is defined as
\begin{equation}
\mathcal{L}(\Theta^{1},...,\Theta^{T}) = \sum\limits_{s=1}^S \ell\left(\mathbf{x}_s^{T}, \mathbf{x}_s^{gt}\right),
\end{equation}
where $T$ is the maximum iteration number. The gradient can be computed by the chain rule,
\begin{equation} \label{eq:SCARL joint gradient}
\frac{\partial \ell(\mathbf{x}^{T},\mathbf{x}^{gt})}{\partial \Theta^{t}} =
\frac{\partial \mathbf{x}^{t}}{\partial \Theta^{t}}
\frac{\partial \mathbf{x}^{t+1}}{\partial \mathbf{x}^{t}}\cdots
\frac{\partial \ell(\mathbf{x}^{T},\mathbf{x}^{gt})}{\partial \mathbf{x}^{T}}.
\end{equation}
%
%
where only $\frac{\partial \mathbf{x}^{t+1}}{\partial \mathbf{x}^{t}}$ need to be additionally computed.
By reformulating the solution in the matrix form,
\begin{equation} \label{eq:supp_SCARL gradient model matrix}
\begin{aligned}
\mathbf{x}^{t+1}&=\mathbf{x}^{t} - \sum\limits_{i=1}^{N_r} ({\mathbf{F}}_i^{t+1})^{\top}\phi^{t+1}_i(\mathbf{F}^{t+1}_i \mathbf{x}^{t})
 \\
-&\lambda^{t+1} \sum\limits_{i=1}^{N_f} \mathcal{A}^{\top} ({\mathbf{P}}_i^{t+1})^{\top}\varphi^{t+1}_i\left({\mathbf{P}}^{t+1}_i \left(\mathcal{A}\mathbf{x}^{t}-\mathbf{y}\right)\right),
\end{aligned}
\end{equation}
the gradient can be computed as
\begin{equation}
\frac{\partial \mathbf{x}^{t+1}}{\partial \mathbf{x}^{t}}
= \mathbf{I} - \sum\limits_{i=1}^{N_r} (\mathbf{F}_i^{t+1})^{\top}\Gamma_i \mathbf{F}_i^{t+1}
- \sum\limits_{i=1}^{N_f} \mathcal{A}^{\top} (\mathbf{P}_i^{t+1})^{\top}\Lambda_i \mathbf{P}_i^{t+1}\mathcal{A},
\end{equation}
where $\Gamma_i =\text{diag} \left(\phi^{t'}_i(z_1),...,\phi^{t'}_i(z_N)\right)$ is also a diagonal matrix.

Once $\frac{\partial \mathbf{x}^{t+1}}{\partial \mathbf{x}^{t}}$ is computed, the overall gradient can be computed by the chain rule and the other gradient parts in \eqref{eq:SCARL joint gradient} can be borrowed from greedy training.

%
%

\vspace{-.08in}
\subsubsection{Training Procedure}

Given a training dataset, the training of SFARL is to sequentially run greedy training as Algorithm \ref{algm:greedy training} and joint fine-tuning as Algorithm \ref{algm:joint training}.
Algorithm \ref{algm:sfarl inference} lists the inference of SFARL given model parameters, in which all the intermediate results are recorded for backward propagation during training.
In greedy training $\Theta^{t+1}$, parameters $\{\Theta^i\}_{i=1}^{t}$ in previous $t$ stages are fixed, and only gradients in stage $t+1$ are computed and are fed to ADAM algorithm.
In joint fine-tuning, gradients in each stage are computed, and are fed to ADAM algorithm to optimize the parameters $\{\Theta^{t}\}_{t=1}^{T}$ for all the stages.


\vspace{-.08in}
\section{Experimental Results}
\label{sec:experiments}
In this section, we evaluate the proposed SFARL algorithm on several restoration tasks, i.e., image deconvolution either with an inaccurate blur kernel or with multiple degradations, rain streak removal from a single image.
SFARL can also be evaluated on Gaussian denoising, and we have presented the results in the supplementary material. 
%
In our experiments, $7  \times 7$ filters are adopted in both fidelity and regularization terms.
As for stage number, we recommend to set it based on the convergence behavior during greedy training, and empirically use 10-stage SFARL for image deconvolution, and 5-stage SFARL for rain streak removal and Gaussian denoising.
During training SFARL, greedy training ends with 10 epoches for each stage, and then the parameters are further jointly fine-tuned with 50 epochs.
We use ADAM~\cite{kingma2014adam} to optimize these SFARL models with learning rate $1\times 10^{-3}$, $\beta_1 = 0.9$ and $\beta_2 = 0.99$.
Using rain streak removal as an example, it takes about 19 hours to train a SFARL model on a computer equipped with a GTX 1080Ti GPU.
The SFARL models are quantitatively and qualitatively evaluated and compared with state-of-the-art conventional and deep CNN-based approaches.

%
More experimental settings and results are included in the supplementary material.
The source code is available at \url{https://github.com/csdwren/sfarl}.

\vspace{-.1in}
\subsection{Deconvolution with Inaccurate Blur Kernels}


We consider the blind deconvolution task
and use two blur kernel estimation methods, i.e., Cho and  Lee \cite{cho2009fast} and Xu and Jia \cite{xu2010two}, for experiments.
For each estimation approach, we evaluate the performance of SFARL for handling approach-specific blur kernel estimation error.
To construct the training dataset, we use eight blur kernels \cite{levin2009understanding} on 200 clean images from the BSD dataset \cite{martin2001bsd}.
The Gaussian noise with $\sigma=0.25$ is added to generate the blurry images.
The methods by Cho and Lee \cite{cho2009fast} and Xu and  Jia \cite{xu2010two} are used to estimate blur kernels.
Thus, we have 1,600 training samples for each blur kernel estimation approach.
To ensure the training sample quality, we randomly select 500 samples with error ratio \cite{levin2009understanding} above 3 for each image deconvolution method.

%
%
%

\begin{table}[h]\footnotesize
\setlength{\abovecaptionskip}{1pt}
\setlength{\belowcaptionskip}{0pt}
\centering
\setlength{\tabcolsep}{2pt}
\caption{Quantitative SSIM results on the dataset by Levin et al. \cite{levin2009understanding}.
} \label{table:deblur ssim levin}
\begin{tabular}{ccccccc}
\hline

\hline
Kernel estimation  &	EPLL\cite{zoran2011learning}	& ROBUST\cite{ji2012robust} & IRCNN\cite{zhang2017learning} & SFARL\\
\hline
\hline
{Cho and  Lee} \cite{cho2009fast} & 0.8801	& 0.8659 & 0.8825 &0.8903 \\
{Xu and  Jia} \cite{xu2010two}&0.9000	& 0.8917 & 0.9023 &0.9164\\

\hline

\hline
\end{tabular}
\end{table}

\begin{figure*}[!htb]\footnotesize
	\setlength{\abovecaptionskip}{1pt}
	\setlength{\belowcaptionskip}{0pt}
\centering
\setlength{\tabcolsep}{1pt}
\begin{tabular}{ccccccc}
 \includegraphics[width=.16\textwidth]{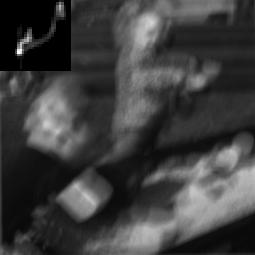} &
 \includegraphics[width=.16\textwidth]{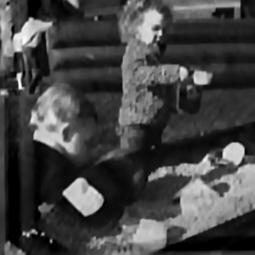} &
 \includegraphics[width=.16\textwidth]{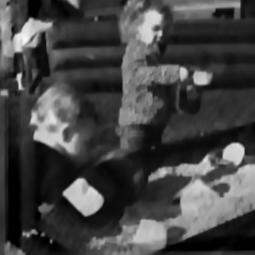}&
 \includegraphics[width=.16\textwidth]{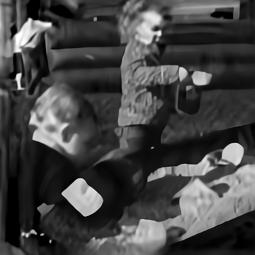}&
 \includegraphics[width=.16\textwidth]{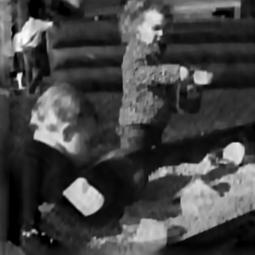}\\
 Blurry image &  EPLL \cite{zoran2011learning}& ROBUST \cite{ji2012robust}& IRCNN \cite{zhang2017learning}& SFARL \\
\end{tabular}
   \caption{Visual quality comparison on Levin et al.'s dataset \cite{levin2009understanding}. }
\label{fig:sfarl deblur}
\end{figure*}

\begin{figure*}[!htb]\footnotesize
\setlength{\abovecaptionskip}{1pt}
\setlength{\belowcaptionskip}{0pt}
\centering
\begin{tabular}{cclcclcclccl}
   \multicolumn{3}{c}{\includegraphics[width=.21\textwidth]{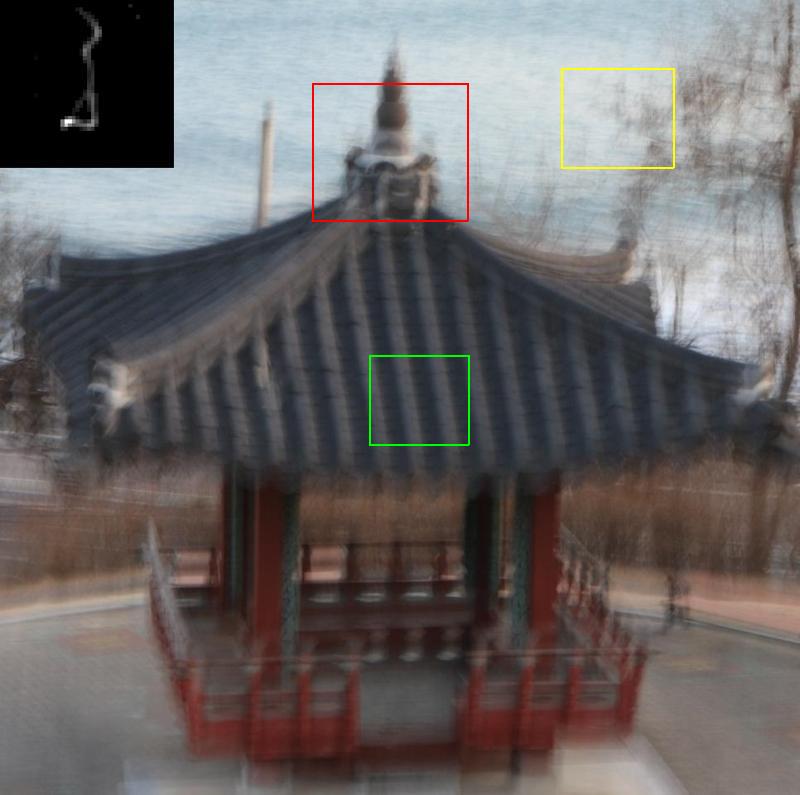}}\ &
   \multicolumn{3}{c}{\includegraphics[width=.21\textwidth]{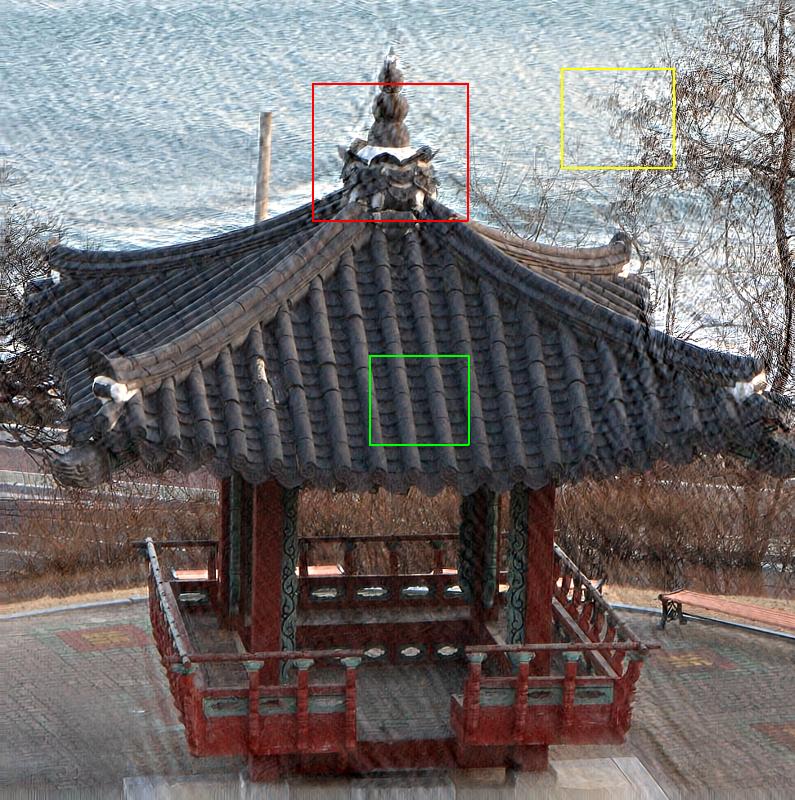}}\ &
   \multicolumn{3}{c}{\includegraphics[width=.21\textwidth]{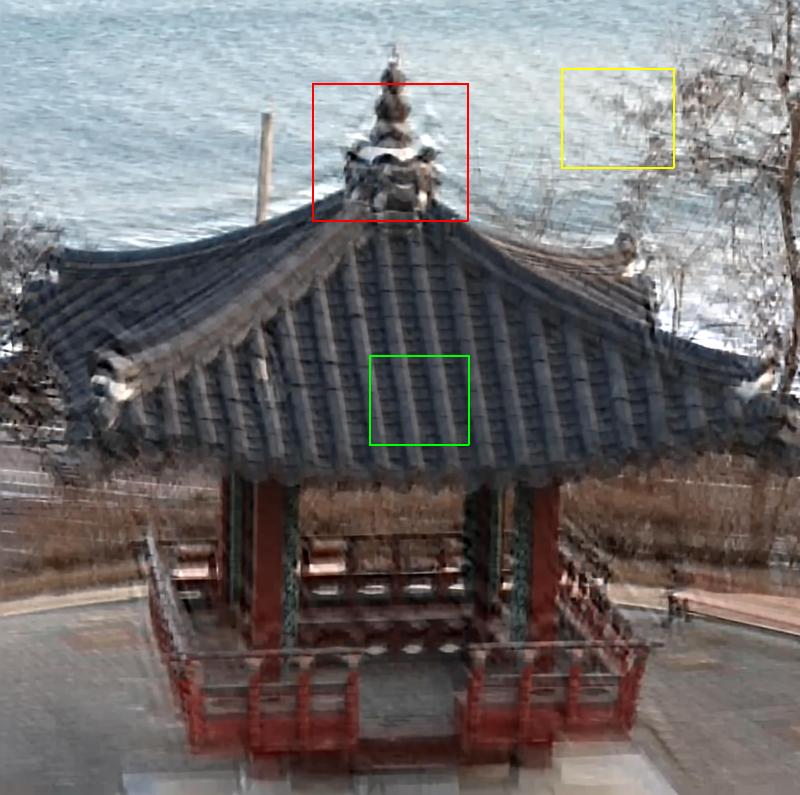}}\ &
   \multicolumn{3}{c}{\includegraphics[width=.21\textwidth]{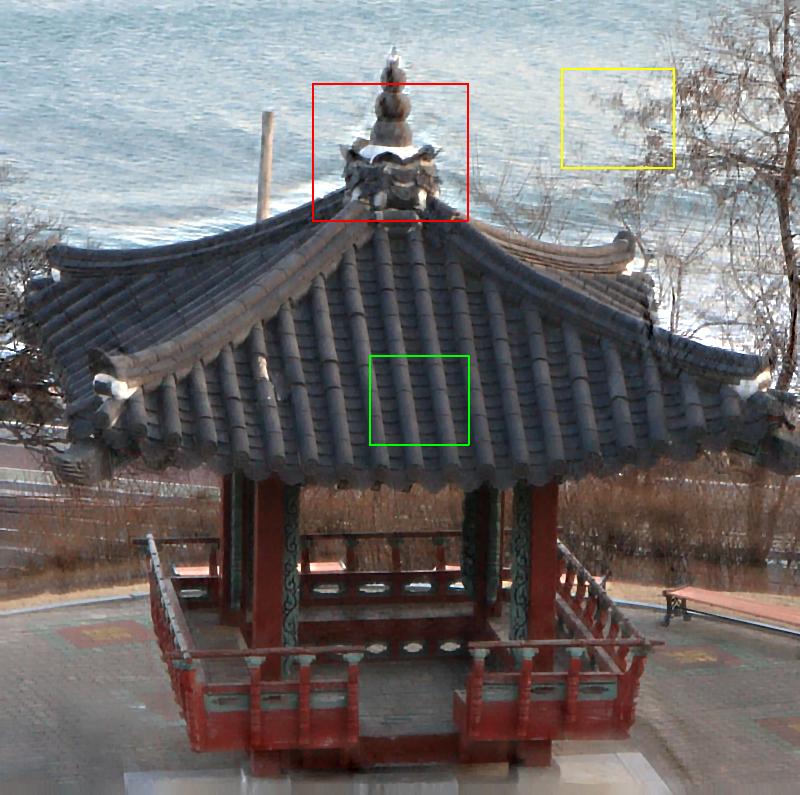}}\vspace{-2pt}\\
   \includegraphics[width=.07\textwidth]{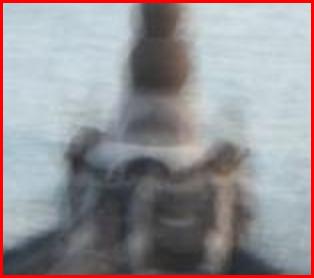}&
   \includegraphics[width=.07\textwidth]{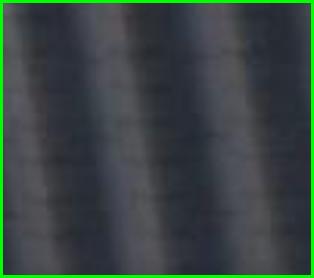}&
   \includegraphics[width=.07\textwidth]{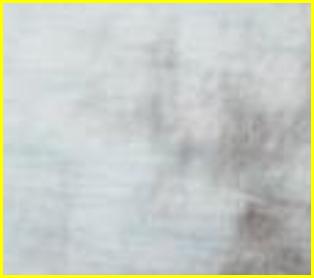}\ &
   \includegraphics[width=.07\textwidth]{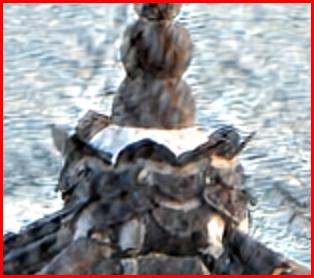}&
   \includegraphics[width=.07\textwidth]{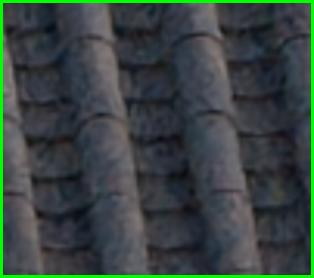}&
   \includegraphics[width=.07\textwidth]{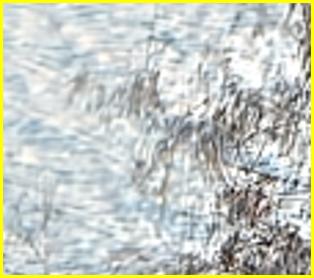}\ &
   \includegraphics[width=.07\textwidth]{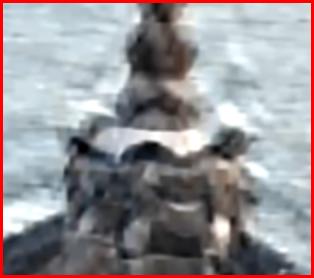}&
   \includegraphics[width=.07\textwidth]{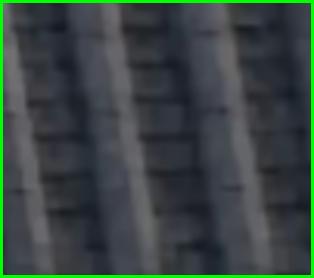}&
   \includegraphics[width=.07\textwidth]{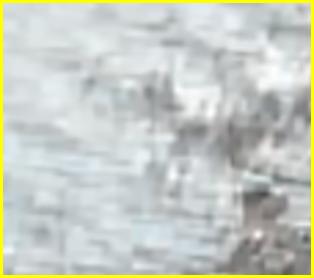}\ &
   \includegraphics[width=.07\textwidth]{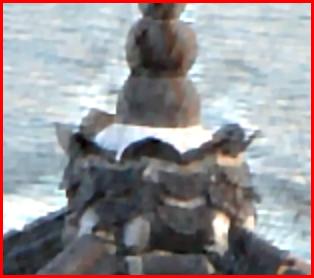}&
   \includegraphics[width=.07\textwidth]{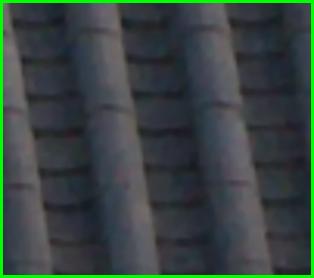}&
   \includegraphics[width=.07\textwidth]{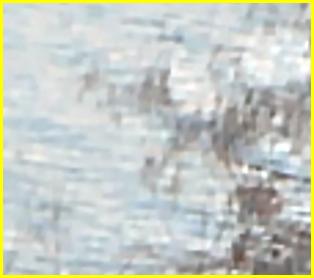}\\
\multicolumn{3}{c}{Blurry images} &
\multicolumn{3}{c}{IRCNN \cite{zhang2017learning}} &
\multicolumn{3}{c}{ROBUST \cite{ji2012robust}} &
\multicolumn{3}{c}{SFARL}\\

\end{tabular}
\caption{Deblurring results on real blurry images, in which blur kernels are estimated by Xu and  Jia \cite{xu2010two}.}
\label{fig:deconvolution real results}
\end{figure*}



 On the widely used synthetic dataset, i.e., Levin et al. \cite{levin2009understanding}, we compare our SFARL with
EPLL \cite{zoran2011learning}, ROBUST \cite{ji2012robust} and IRCNN \cite{zhang2017learning}.
%
%
The testing dataset includes 4 clean images and 8 blur kernels.
The blur kernels are estimated by Cho and  Lee \cite{cho2009fast} and Xu and  Jia \cite{xu2010two}.
Table \ref{table:deblur ssim levin}
lists the average SSIM values of all evaluated methods
on the dataset by Levin et al. \cite{levin2009understanding}.
Overall, the SFARL algorithm performs favorably against the other methods
 in terms of SSIM.
%
%
From Table \ref{table:deblur ssim levin}, we also have the following observations.
First, the SFARL algorithm models the residual images by specific blur kernel estimation method to improve restoration result.
For each blur kernel estimation method, what we need to do is to retrain the SFARL model from the synthetic data.
Second, when the estimated blur kernel is more accurate (e.g., Xu and  Jia \cite{xu2010two}), better quantitative performance indexes are also attained by our SFARL.
%

%


We evaluate the SFARL algorithm against the state-of-the-art methods
on a synthetic and a real blurry images in Figures \ref{fig:sfarl deblur} and \ref{fig:deconvolution real results}.
The blur kernels are estimated using the method by Xu and  Jia \cite{xu2010two}.
As the blur kernel can be accurately estimated in Fig. \ref{fig:sfarl deblur},
all the evaluated methods perform well and the SFARL algorithm
restores more texture details.
On the other hand, the estimated blur kernel is less accurately estimated in Fig. \ref{fig:deconvolution real results}.
Among all the evaluated methods, the deblurred image by the SFARL algorithm is sharper with
fewer ringing effects than those by the other methods.
We note that  IRCNN \cite{zhang2017learning} use the $\ell_2$-norm in the fidelity term
 and the ROBUST scheme \cite{ji2012robust} introduces an $\ell_1$-norm regularizer
on the residual $\mathbf{z}$ caused by kernel error.
However, both $\ell_2$-norm and $\ell_1$-norm are limited in modeling the complex distribution of the residual, and neither GMM prior in EPLL nor deep CNN prior in IRCNN cannot well compensate the effect caused by inaccurate blur kernels.
%
Thus, the performance gain of the SFARL model can be attributed to its effectiveness in characterizing the spatial dependency and complex distribution of residual images.

\vspace{-.1in}
\subsection{Deconvolution with Multiple Degradations }
We consider a more challenging deconvolution task \cite{xu2014deep}, in which blur convolution is followed by multiple degradations including saturation, Gaussian noise and JPEG compression.
SFARL is compared with DCNN \cite{xu2014deep}, Whyte \cite{whyte2014deblurring}, IRCNN \cite{zhang2017learning} and SRN \cite{tao2018scale}.
Following the degradation steps in \cite{xu2014deep}, 500 clean images from BSD dataset \cite{martin2001bsd} are used to synthesize training dataset, on which SFARL and SRN are trained.
Since only testing code of DCNN \cite{xu2014deep} and 30 testing images on a disk kernel with radius 7 (Disk7) are released, SFARL is only evaluated on Disk7 kernel.
From Table \ref{table:sfarl deblur disk},
SFARL performs favorably in terms of average PSNR and SSIM.
The results by SFARL are also visually more pleasing, while the results by the other methods suffer from visible noises and artifacts, as shown in Fig. \ref{fig:sfarl deblur disk}.
It is worth noting that IRCNN works well in reducing blurring, but magnifies other degradations to yield ringing effects and noises.
SRN is an up-to-date deep motion deblurring network, but is still suffering from visible noises and artifacts, since the ill-poseness caused by disk blur is usually more severe than motion blur.
Thus, we conclude that SFARL is able to model these multiple degradations in fidelity term.
Moreover, it should be noted that DCNN needs to initialize deconvolution sub-network using inverse kernels, while our SFARL is much easier to train given proper training dataset.
\begin{table}[!htbp] \footnotesize
	\setlength{\abovecaptionskip}{1pt}
	\setlength{\belowcaptionskip}{0pt}
\setlength{\tabcolsep}{4pt}
\centering
\caption{Quantitative comparison on deconvolution with multiple degradations \cite{xu2014deep}.} \label{table:sfarl deblur disk}
\begin{tabular}{ccccccccc}
\hline

\hline
Method & Whyte\cite{whyte2014deblurring} & DCNN\cite{xu2014deep} & IRCNN\cite{zhang2017learning} & SRN\cite{tao2018scale}& SFARL \\
\hline
\hline
PSNR &26.35 & 26.50 & 23.84 & 26.46&  26.66   \\
SSIM & 0.8307& 0.8442& 0.6673& 0.8447&  0.8532\\
\hline

\hline
\end{tabular}
\end{table}
\begin{figure*}[!htb]\footnotesize
	\setlength{\abovecaptionskip}{1pt}
	\setlength{\belowcaptionskip}{0pt}
\centering
\setlength{\tabcolsep}{1pt}
\begin{tabular}{ccccccc}
 \includegraphics[width=.24\textwidth]{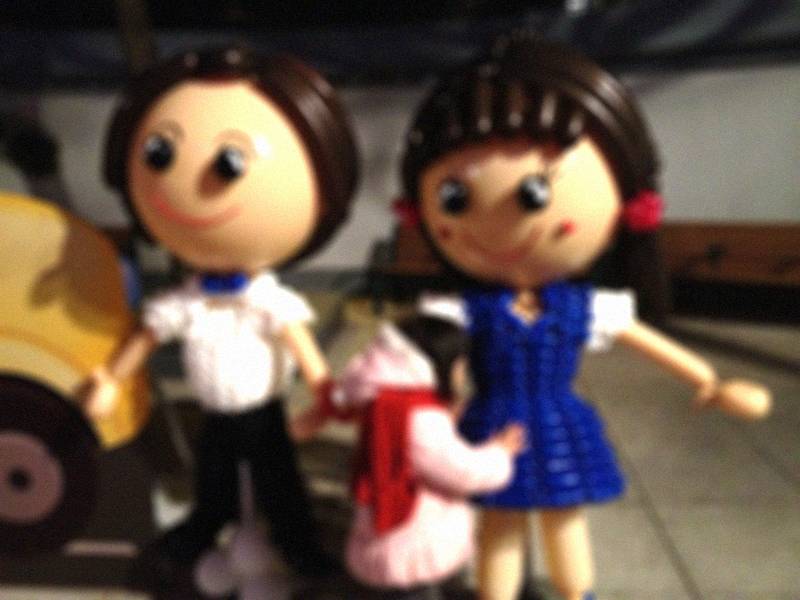}  &
 \includegraphics[width=.24\textwidth]{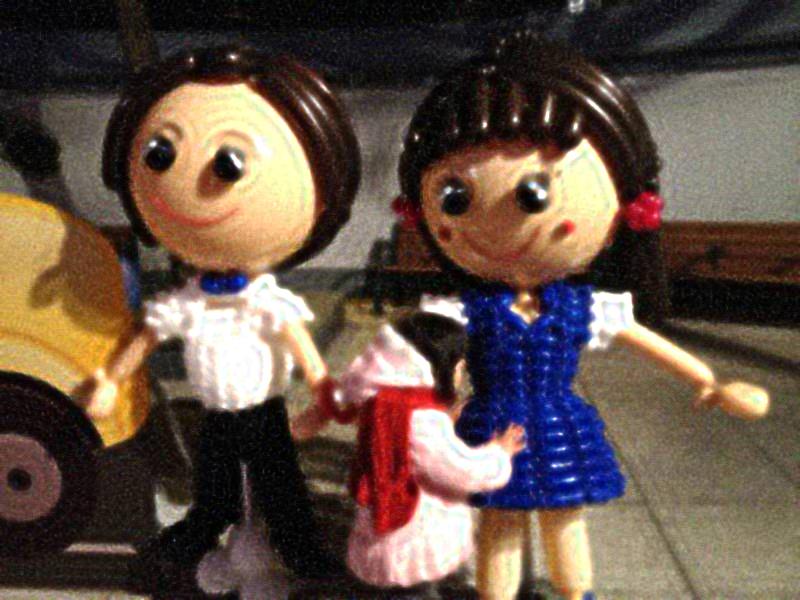} &
 \includegraphics[width=.24\textwidth]{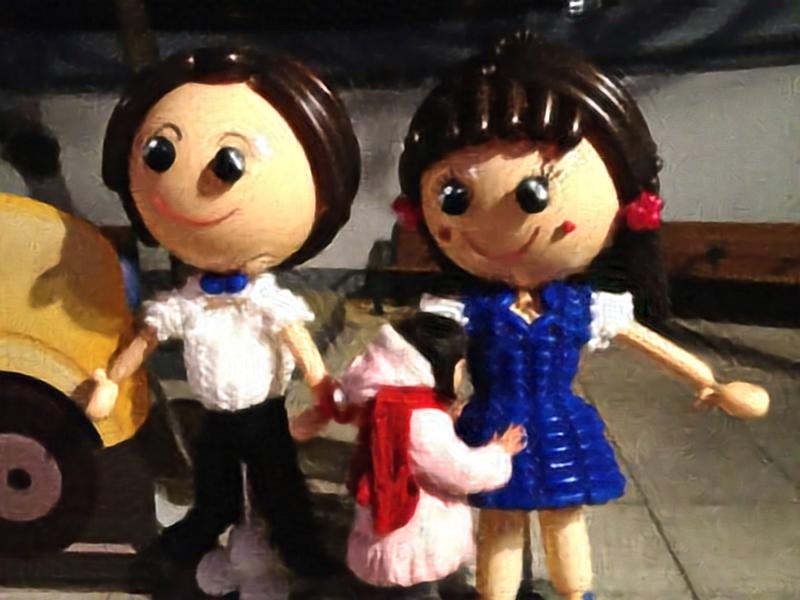} \\
 Blurry image & Whyte \cite{whyte2014deblurring} & DCNN \cite{xu2014deep} \\
 \includegraphics[width=.24\textwidth]{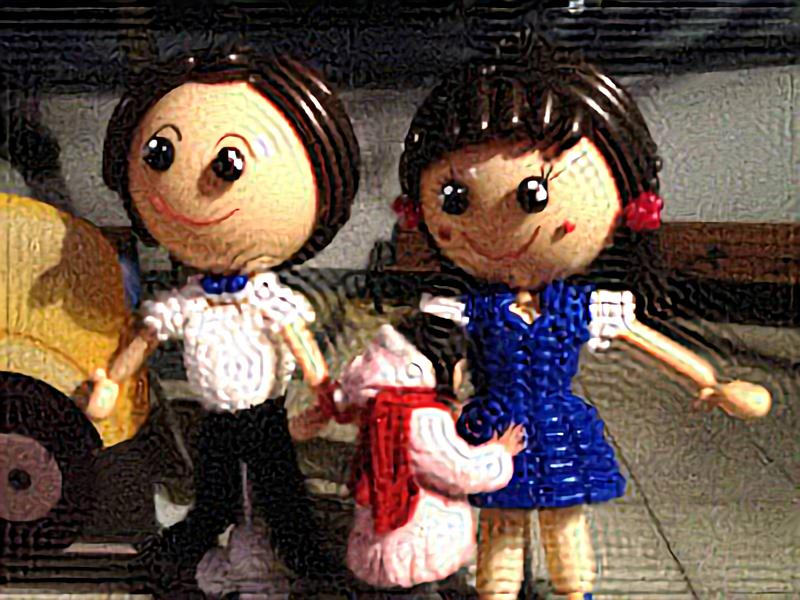}  &
 \includegraphics[width=.24\textwidth]{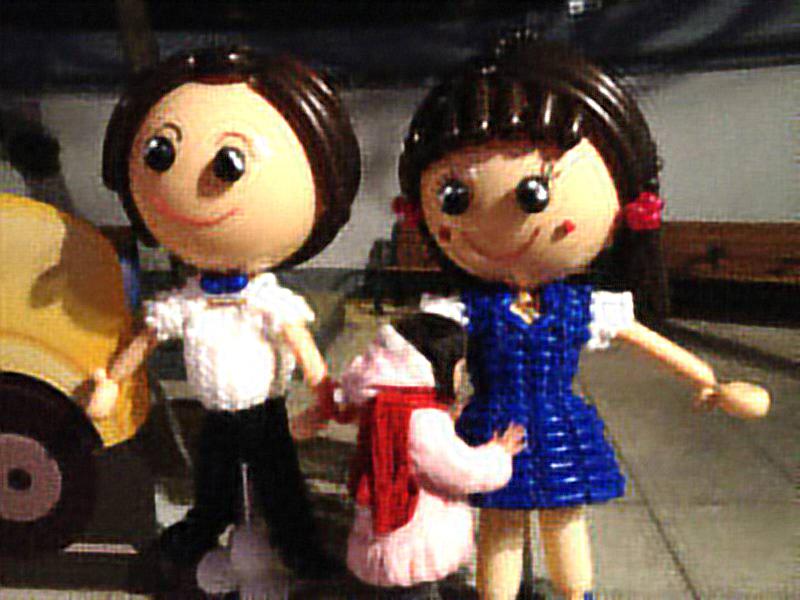}  &
 \includegraphics[width=.24\textwidth]{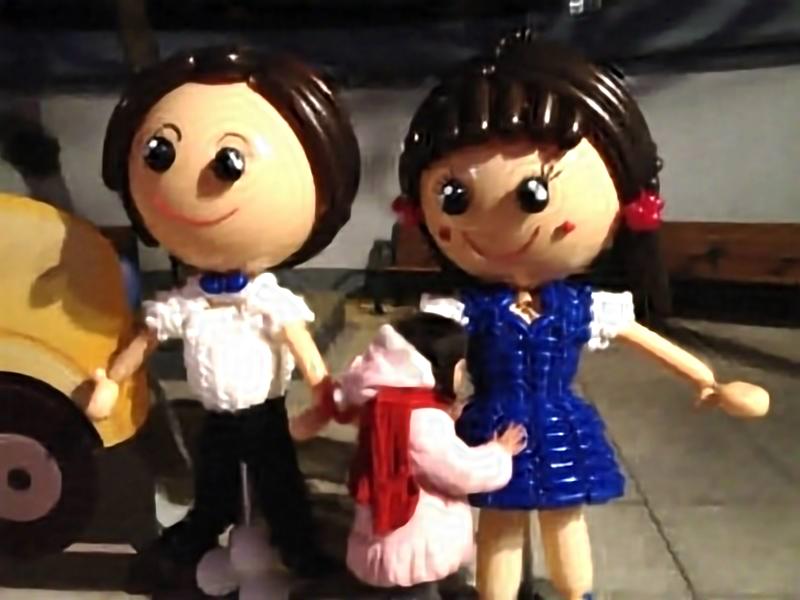} \\
 IRCNN \cite{zhang2017learning}&SRN \cite{tao2018scale} & SFARL \\
\end{tabular}
   \caption{Visual quality comparison on deconvolution along with Gaussian noise, satature and JPEG compression. }
\label{fig:sfarl deblur disk}
\end{figure*}

\vspace{-.1in}
\subsection{Singe Image Rain Streak Removal} \label{sec:experiment rain}
\begin{figure*}[!htb]\footnotesize
\setlength{\abovecaptionskip}{1pt}
\setlength{\belowcaptionskip}{0pt}
\centering

\begin{tabular}{ccc}
  \includegraphics[width=.21\textwidth]{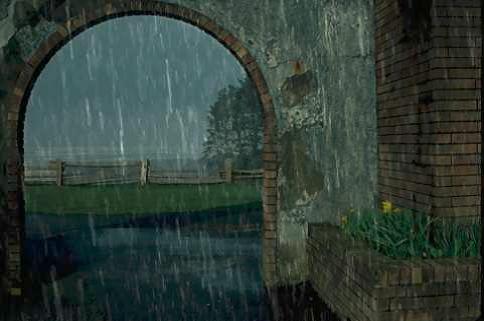} \ &
  \includegraphics[width=.21\textwidth]{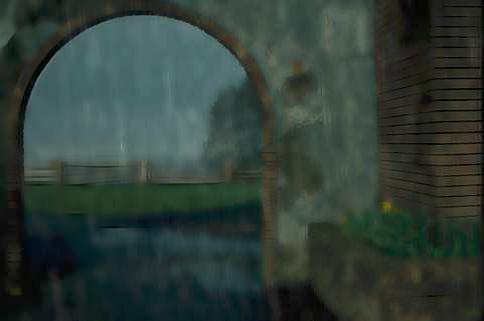} \ &
  \includegraphics[width=.21\textwidth]{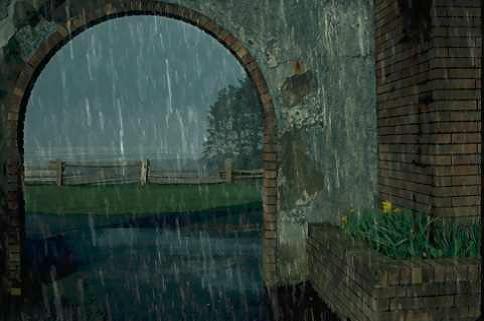} \\
  Rainy image  & SR \cite{kang2012automatic} & LRA \cite{chen2013generalized} \\
  \includegraphics[width=.21\textwidth]{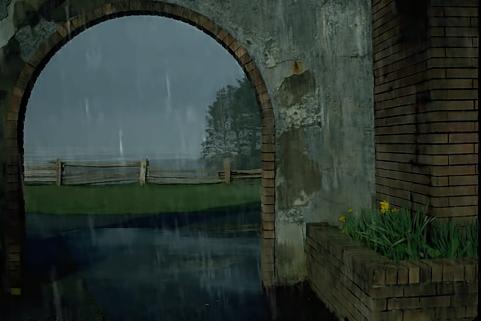} \  &
  \includegraphics[width=.21\textwidth]{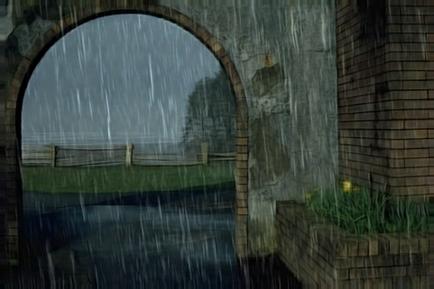} \ &
  \includegraphics[width=.21\textwidth]{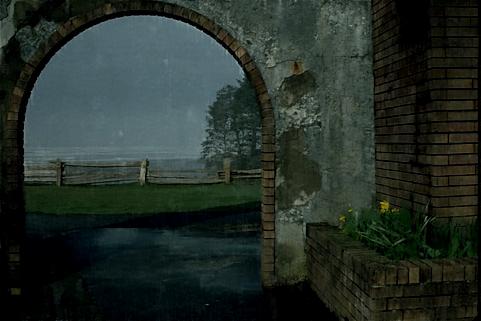} \\
  GMM \cite{li2016rain} & CNN \cite{eigen2013restoring} & SFARL
\end{tabular}
\caption{Rain streak removal results of five evaluated methods on a synthetic image in \cite{li2016rain}.}
\label{fig:rain synthetic_supp}
\end{figure*}

To train the SFARL model for rain streak removal, we construct a synthetic rainy dataset.
We randomly select 100 clean outdoor images from the UCID dataset \cite{schaefer2003ucid}, and use the Photoshop function (\url{http://www.photoshopessentials.com/photo-effects/rain/})
to generate 7 rainy images at 7 random rain scales and different orientations ranged from 60 to 90 degrees.
The training dataset contains 700 images with different rain orientations and scales.

\begin{table*}[!htbt]\footnotesize
\setlength{\abovecaptionskip}{1pt}
\setlength{\belowcaptionskip}{0pt}
\centering
\setlength{\tabcolsep}{6pt}
\caption{Deraining results on synthetic rainy images in \cite{li2016rain} in terms of SSIM} \label{table:rain ssim}
\begin{tabular}{cccccccccccccc}
\hline

\hline
Method & \#1 & \#2 & \#3 & \#4 & \#5 & \#6 & \#7 & \#8 & \#9 & \#10 & \#11 & \#12 & Avg.\\
\hline
\hline
SR\cite{kang2012automatic} &0.74& 0.79& 0.84& 0.77& 0.63& 0.73 &0.82& 0.77 &0.74& 0.74 &0.65& 0.77 &0.75\\
LRA\cite{chen2013generalized} &0.83 &0.88 &0.76 &0.96 &0.92 &0.93 &0.94 &0.81& 0.90 &0.82 &0.85 &0.80 &0.87\\
GMM\cite{li2016rain}& 0.89& 0.93& 0.92& 0.94 &0.90& 0.95& 0.96& 0.90 &0.91 &0.90 &0.86& 0.92& 0.91\\
CNN\cite{eigen2013restoring} & 0.75 &0.79 &0.71 &0.89 &0.76& 0.80 &0.85& 0.77 &0.81& 0.76 &0.79& 0.73 &0.78\\
SFARL & 0.93  &  0.93  &  0.92  &  0.95   & 0.97   & 0.94& 0.98   & 0.95   & 0.97 &   0.98 &   0.95  &  0.97& 0.95\\
\hline

\hline
\end{tabular}
\end{table*}

We evaluate the SFARL method with the state-of-the-art algorithms including SR \cite{kang2012automatic}, LRA \cite{chen2013generalized}, GMM \cite{li2016rain}, and the CNN \cite{eigen2013restoring}, on a the synthetic dataset \cite{li2016rain}.
The dataset consists of 12 rainy images with orientation ranged from left to right.
%
%
 Table \ref{table:rain ssim}
shows that
the SFARL algorithm achieves the highest SSIM values for each test image.
Fig. \ref{fig:rain synthetic_supp} shows
rain streak removal results by all the evaluated algorithms on a synthetic rainy image.
The results by the SFARL and GMM algorithms are significantly better than the other methods.
However, the result by the GMM method still has visible rain streaks, while the SFARL model recovers satisfying clean image.
%
%

\begin{figure*}[!htb]\footnotesize
	\setlength{\abovecaptionskip}{1pt}
	\setlength{\belowcaptionskip}{0pt}
\centering
\setlength{\tabcolsep}{1pt}
\begin{tabular}{ccccccc}
 \includegraphics[width=.21\textwidth]{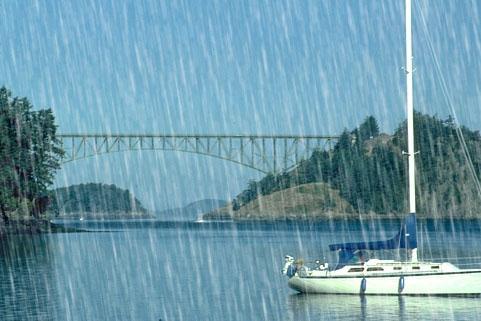}  &
 \includegraphics[width=.21\textwidth]{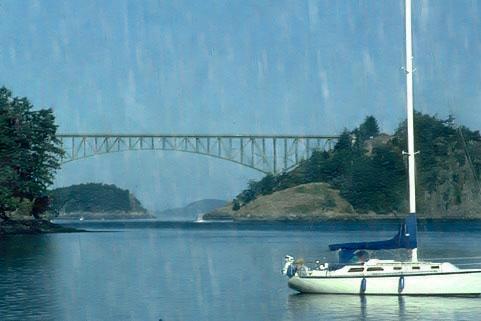}&
 \includegraphics[width=.21\textwidth]{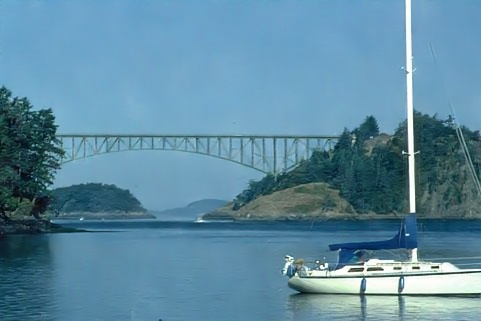} \\
	\includegraphics[width=.21\textwidth]{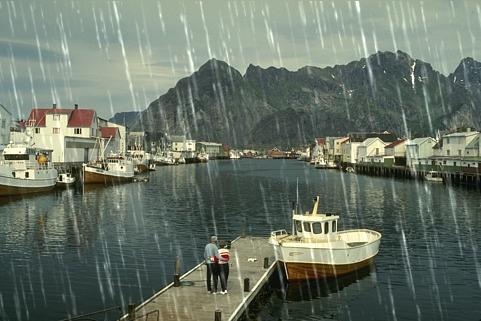}  &
	\includegraphics[width=.21\textwidth]{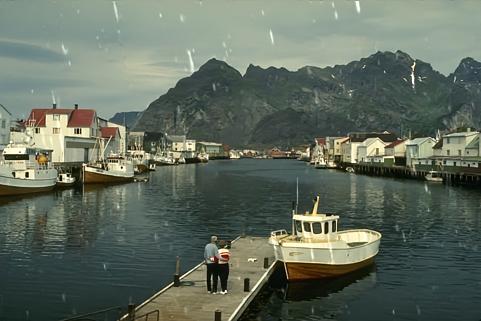}  &
	\includegraphics[width=.21\textwidth]{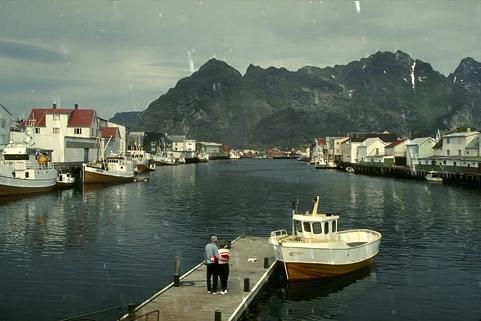}  \\
 Rainy images  &DDNET \cite{fu2017removing}& SFARL \\
\end{tabular}
   \caption{Visual quality comparison for rain streak removal.
   Both SFARL and DDNET are trained for Rain1400 \cite{fu2017removing}.
   	The first row is from Rain1400 \cite{fu2017removing}, while the second row is from Rain100L \cite{yang2017deep} for evaluating generalization ability.   }
\label{fig:sfarl rain ddcnn}
\end{figure*}

Furthermore, we compare SFARL with a recent deep CNN-based method, i.e., DDNET \cite{fu2017removing}.
The authors \cite{fu2017removing} provide a training dataset of 12,600 rainy images and a testing dataset of 1,400 rainy images (Rain1400).
We train SFARL on the training dataset, and on the testing dataset, SFARL is quantitatively and qualitatively compared with DDNET.
%
From Table \ref{table:sfarl derain ddcnn}, SFARL obtains better PSNR and SSIM values on Rain1400.
In Fig. \ref{fig:sfarl rain ddcnn}, SFARL produces satisfactory deraining results, while rain streaks are still visible in the results by DDNET.

\begin{table}[htbp] \footnotesize
	\setlength{\abovecaptionskip}{1pt}
	\setlength{\belowcaptionskip}{0pt}
\centering
\setlength{\tabcolsep}{10pt}
\caption{Average PSNR/SSIM comparison on Rain1400 \cite{fu2017removing} and Rain100L \cite{yang2017deep}. Both SFARL and DDNET are trained for Rain1400, and are directly used to process Rain100L to validate generalization ability. } \label{table:sfarl derain ddcnn}
\begin{tabular}{cccc}
	\hline
	
	\hline
	Method & DDNET \cite{fu2017removing} & SFARL \\
	\hline
	\hline
	Rain1400\cite{fu2017removing} &  29.91/0.9099 & 31.37/0.9188 \\
	Rain100L\cite{yang2017deep} & 29.12/0.9012 & 29.73/0.9181  \\
	\hline
	
	\hline
\end{tabular}
\end{table}

\begin{figure*}[!htb]\footnotesize
\setlength{\abovecaptionskip}{1pt}
\setlength{\belowcaptionskip}{0pt}
\centering
\begin{tabular}{cclcclcclccl}
   \multicolumn{3}{c}{\includegraphics[width=.21\textwidth]{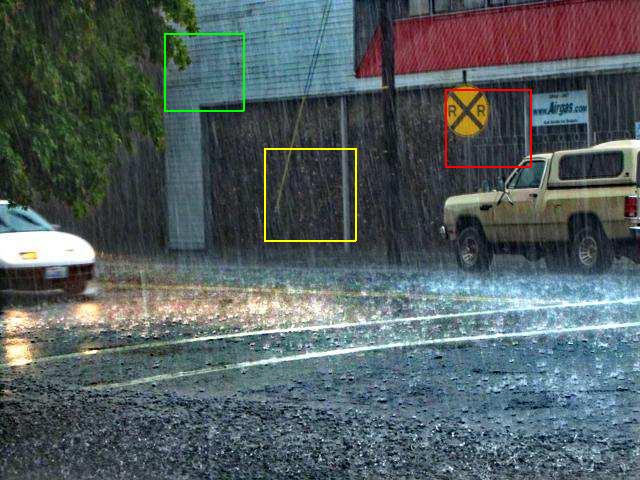}}\ &
   \multicolumn{3}{c}{\includegraphics[width=.21\textwidth]{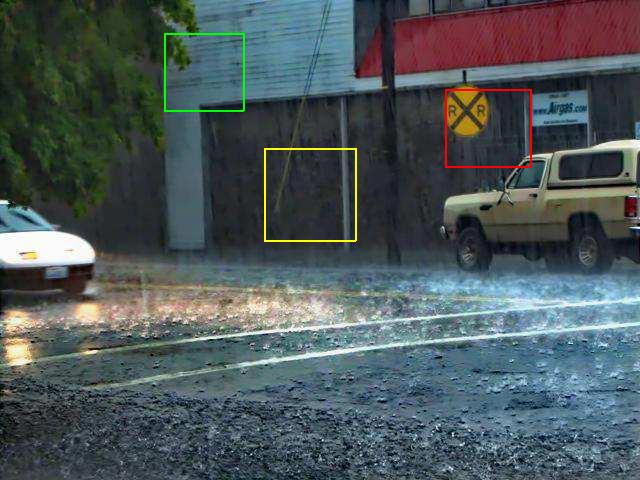}}\ &
   \multicolumn{3}{c}{\includegraphics[width=.21\textwidth]{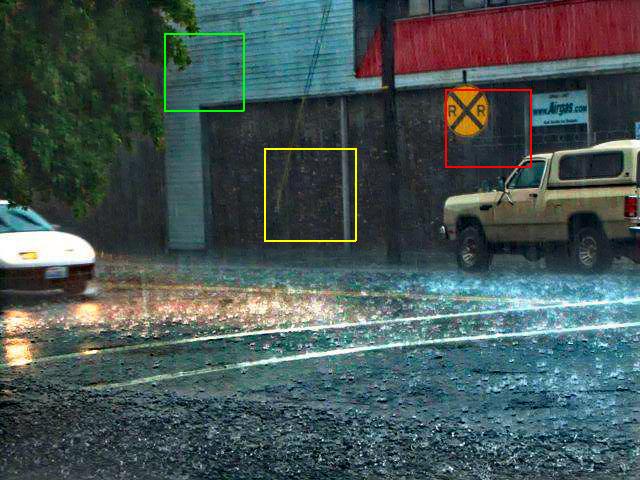}}\ &
   \multicolumn{3}{c}{\includegraphics[width=.21\textwidth]{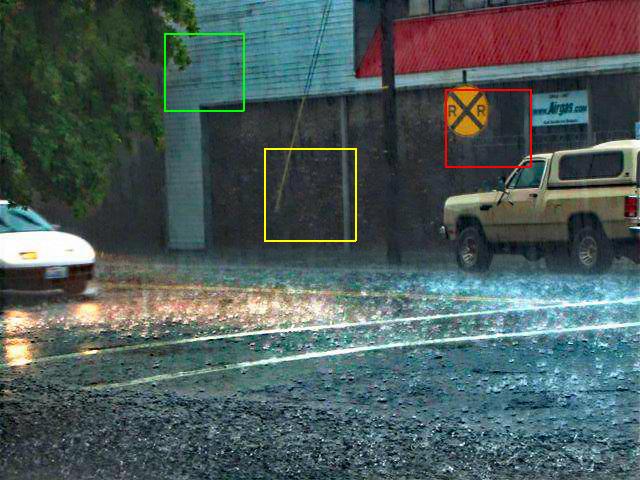}}\vspace{-2pt}\\
   \includegraphics[width=.07\textwidth]{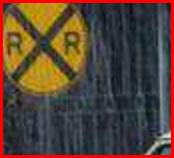}&
   \includegraphics[width=.07\textwidth]{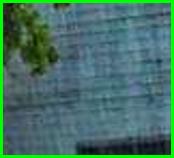}&
   \includegraphics[width=.07\textwidth]{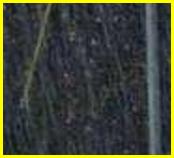}\ &
   \includegraphics[width=.07\textwidth]{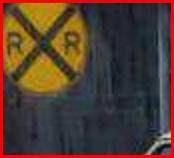}&
   \includegraphics[width=.07\textwidth]{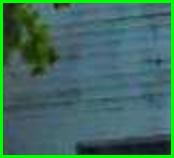}&
   \includegraphics[width=.07\textwidth]{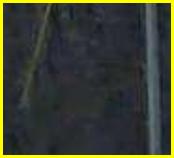}\ &
   \includegraphics[width=.07\textwidth]{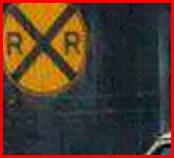}&
   \includegraphics[width=.07\textwidth]{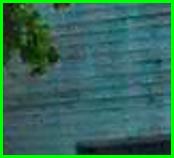}&
   \includegraphics[width=.07\textwidth]{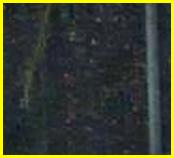}\ &
   \includegraphics[width=.07\textwidth]{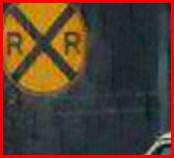}&
   \includegraphics[width=.07\textwidth]{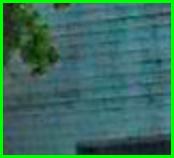}&
   \includegraphics[width=.07\textwidth]{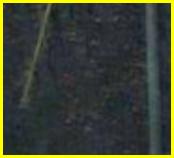}\\
   \multicolumn{3}{c}{\includegraphics[width=.21\textwidth]{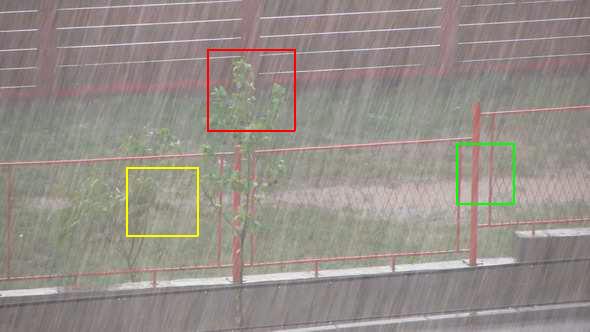}}\ &
   \multicolumn{3}{c}{\includegraphics[width=.21\textwidth]{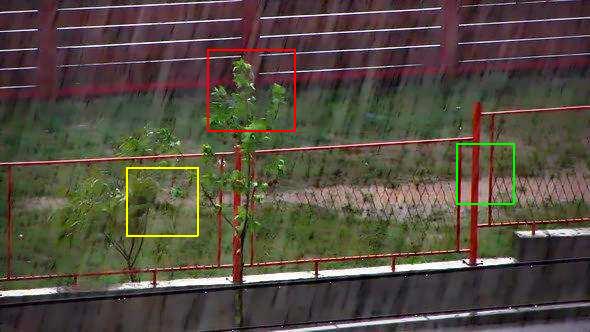}}\ &
    \multicolumn{3}{c}{\includegraphics[width=.21\textwidth]{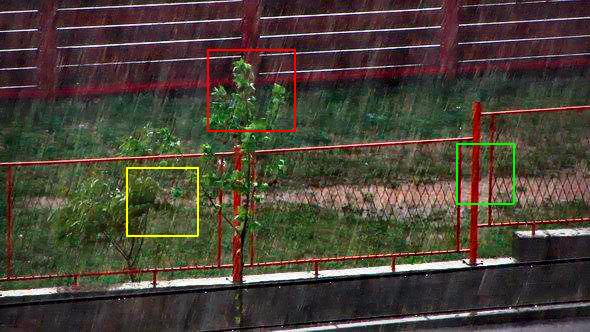}}\ &
    \multicolumn{3}{c}{\includegraphics[width=.21\textwidth]{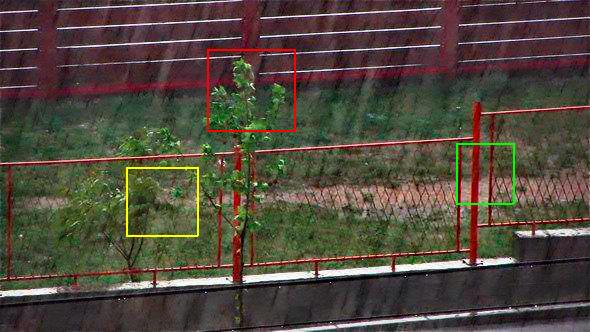}}\vspace{-2pt}\\
    \includegraphics[width=.07\textwidth]{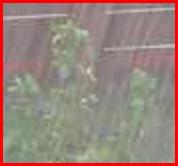}&
    \includegraphics[width=.07\textwidth]{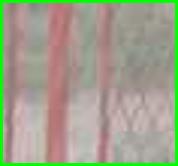}&
    \includegraphics[width=.07\textwidth]{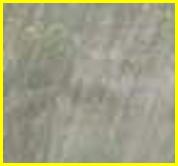}\ &
    \includegraphics[width=.07\textwidth]{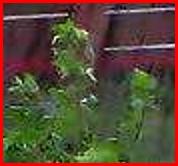}&
    \includegraphics[width=.07\textwidth]{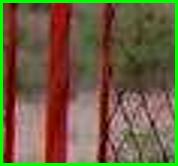}&
    \includegraphics[width=.07\textwidth]{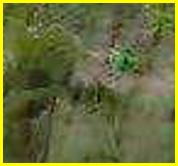}\ &
    \includegraphics[width=.07\textwidth]{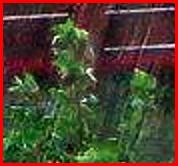}&
    \includegraphics[width=.07\textwidth]{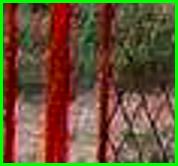}&
    \includegraphics[width=.07\textwidth]{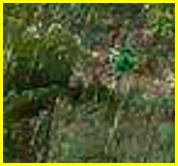}\ &
    \includegraphics[width=.07\textwidth]{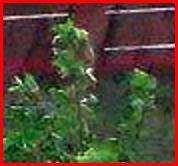}&
    \includegraphics[width=.07\textwidth]{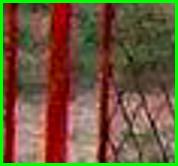}&
    \includegraphics[width=.07\textwidth]{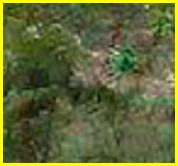}\\
\multicolumn{3}{c}{Rainy images} &
\multicolumn{3}{c}{GMM\cite{li2016rain}} &
\multicolumn{3}{c}{DDNET\cite{fu2017removing}} &
\multicolumn{3}{c}{SFARL}
\end{tabular}
\caption{Results on real rainy images. The rain in second image is very heavy and we first dehaze \cite{meng2013efficient} it to make rain streaks more visible.}
\label{fig:rain real}
\end{figure*}

Moreover, we evaluate the SFARL model on real world rainy images against the state-of-the-art methods.
%
%
Since the rain in second image of Fig. \ref{fig:rain real}
is too heavy to see rain streaks, we first use the dehazing method \cite{meng2013efficient}
before applying a deraining algorithm.
On both test images, the SFARL algorithm performs better than DDNET \cite{fu2017removing} and GMM \cite{li2016rain}.
%
For real rainy images, the image formation process is complex and may not be well characterized by either linear additive model nor screen blend model.
Nevertheless, due to the flexibility of the fidelity term in modeling spatially dependent and highly complex patterns, the SFARL model is more effective in modeling the complex degradation process and achieving satisfactory deraining result.
%
%

\vspace{-.1in}
\subsection{Ablation Study}
In this section, we take rain streak removal as an example to analyze training convergence and effect of negative SSIM loss.
We also evaluate the generalization and transferring ability of SFARL.
Besides, the visualization of learned filters, interpretability and flexibility of fidelity term as well as more discussions on stage number setting are presented in the supplementary material.

\vspace{-.1in}
\subsubsection{Convergence}
\begin{figure}[!htb]\footnotesize
	\setlength{\abovecaptionskip}{1pt}
	\setlength{\belowcaptionskip}{0pt}
\centering
\begin{tabular}{ccccccc}
\hspace{-0.15in}
 \includegraphics[height=.16\textwidth]{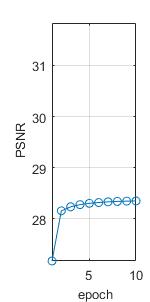} &
 \includegraphics[height=.16\textwidth]{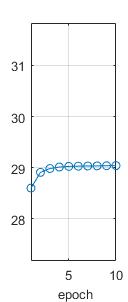} &
 \includegraphics[height=.16\textwidth]{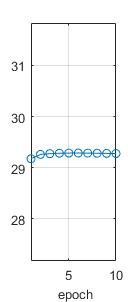} &
 \includegraphics[height=.16\textwidth]{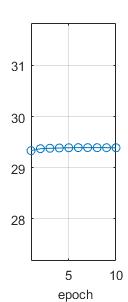} &
 \includegraphics[height=.16\textwidth]{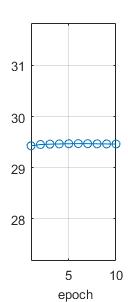} &
 \includegraphics[height=.16\textwidth]{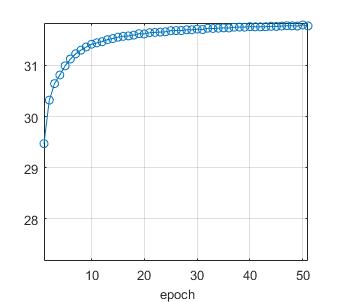} \\
 Stage 1 & Stage 2& Stage 3 & Stage 4 & Stage 5 & Joint Training\\
\end{tabular}
   \caption{Empirical convergence of a 5-stage SFARL for rain streak removal during greedy training and joint fine-tuning. }
\label{fig:ocnvergence}
\end{figure}

As shown in Fig. \ref{fig:ocnvergence}, average PSNR of each epoch is computed to form the converge curves in the 5 stages of greedy training and the final joint fine-tuning.
In greedy training, SFARL can stably converge in every stage, in which
notable performance gains can be attained in the first two stages, while the PSNR increases marginally in the last 3 stages.
After greedy training, SFARL is further jointly fine-tuned, and empirically
converge to a much better solution.

\vspace{-.1in}
\subsubsection{Loss function}
To verify the effect of negative SSIM loss, we train two SFARL models, which share the same settings except training loss, i.e., one is trained by minimizing MSE loss (SFARL-MSE), while the other one by minimizing negative SSIM loss.
These two SFARL models are trained and tested on the datasets provided by \cite{fu2017removing}.
Form Table \ref{table:mse vs ssim}, it is reasonable to see that SFARL-MSE leads to a higher average PSNR value, while SFARL-SSIM performs better in terms of SSIM metric.
Moreover, SFARL-SSIM can better remove rain streaks than SFARL-MSE, e.g., \emph{sky} region in Fig. \ref{fig:mse vs ssim}, indicating that negative SSIM loss is effective in attaining result with higher visual quality.
\begin{table}[!htb] \footnotesize
\setlength{\abovecaptionskip}{1pt}
\setlength{\belowcaptionskip}{0pt}
\centering
\setlength{\tabcolsep}{5pt}
\caption{Average PSNR and SSIM on testing dataset \cite{fu2017removing} of SFARL models for rain streak removal trained by MSE loss and SSIM loss} \label{table:mse vs ssim}
\begin{tabular}{ccccc}
\hline

\hline
Training Loss &	SFARL-MSE	& SFARL-SSIM \\
\hline
\hline
{PSNR}	&31.48	& 31.37 \\
{SSIM} & 0.9153	& 0.9188 \\

\hline

\hline
\end{tabular}
\end{table}
\begin{figure*}[!htb]\footnotesize
	\setlength{\abovecaptionskip}{1pt}
	\setlength{\belowcaptionskip}{0pt}
\centering
\begin{tabular}{cccccccccccc}

    \multicolumn{3}{c}{\includegraphics[width=.21\textwidth]{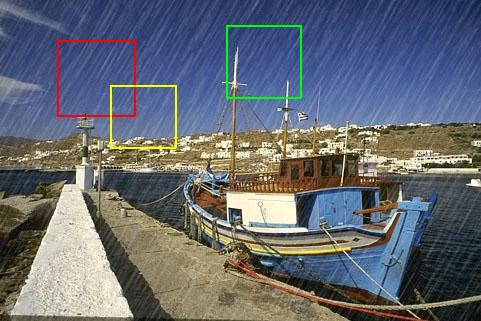}}\ &
   \multicolumn{3}{c}{\includegraphics[width=.21\textwidth]{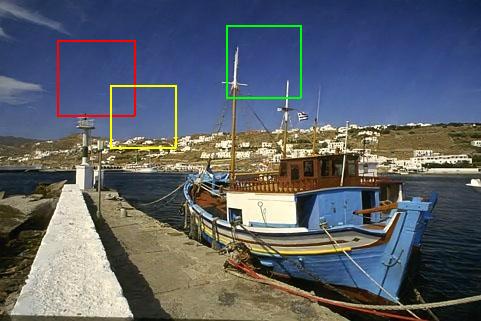}}\ &
   \multicolumn{3}{c}{\includegraphics[width=.21\textwidth]{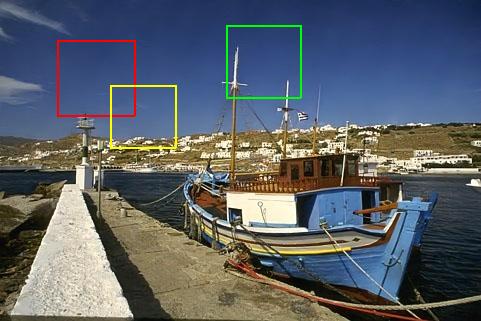}}\vspace{-2pt}\\
   \includegraphics[width=.07\textwidth]{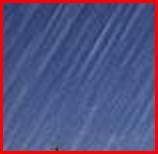}&
   \includegraphics[width=.07\textwidth]{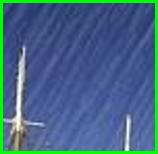}&
   \includegraphics[width=.07\textwidth]{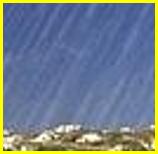}\ &
   \includegraphics[width=.07\textwidth]{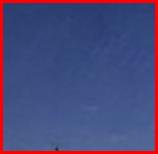}&
   \includegraphics[width=.07\textwidth]{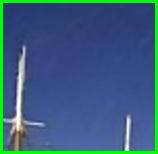}&
   \includegraphics[width=.07\textwidth]{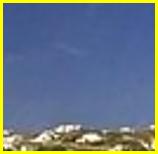}\ &
   \includegraphics[width=.07\textwidth]{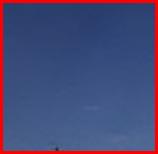}&
   \includegraphics[width=.07\textwidth]{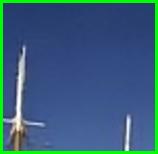}&
   \includegraphics[width=.07\textwidth]{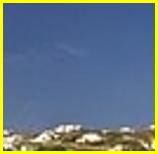}\\
\multicolumn{3}{c}{Rainy image} &
\multicolumn{3}{c}{SFARL-MSE} &
\multicolumn{3}{c}{SFARL-SSIM} \\

\end{tabular}
   \caption{Visual quality comparison of SFARL trained by MSE loss and negative SSIM loss.}
\label{fig:mse vs ssim}
\end{figure*}

\vspace{-.1in}
\subsubsection{Generalization evaluation}
We use the trained model of SFARL for Rain1400 \cite{fu2017removing} in Section \ref{sec:experiment rain} to directly process rainy images in another dataset Rain100L \cite{yang2017deep} for evaluating the generalization ability, and compare it with the deep deraining method DDNET \cite{fu2017removing}.
The rain streaks in Rain1400 and Rain100L are quite different, where those in Rain1400 are dense but gentle, and those in Rain100L are sparse but bright.
And Fig. \ref{fig:sfarl rain ddcnn} shows two rainy images for an intuitive illustration.
From Table \ref{table:sfarl derain ddcnn}, our SFARL can be well generalized to Rain100L.
As shown in Fig.  \ref{fig:sfarl rain ddcnn}, remaining bright rain streaks on Rain100L can still be observed from the results by SFARL and DDNET, indicating that the learning-based methods are limited in handling the cases that are very different from training samples.
Even though, our SFARL exhibits satisfying generalization ability, and there are less remaining rain streaks in the deraining results.


\vspace{-.08in}
\subsubsection{Transferring filters across different tasks}
We discuss the transferring ability of SFARL by applying the learned fidelity and regularization filters across different restoration tasks.
In particular, the filters learned for deconvoluton and denoising tasks are applied to rain streak removal on the Rain12 dataset.
For both denoising and deraining, there are 5 stages for the learned SFARL models, and thus the filters can be transferred in a stage-to-stage manner.
As for deconvolution, the learned SFARL model has 10 stages.
Considering that the first 5 stages are more correlated with deblurring, we apply the filters in the last 5 stages to rain streak removal.
In the following, we respectively discuss the transferring ability of fidelity and regularization filters.

First, we apply the regularization filters learned for deconvolution and denoising to the SFARL model for deraining, denoted by SFARL$_{\text{BlurReg}}$ and SFARL$_{\text{NoiseReg}}$, respectively.
From Table \ref{table:filters transfer}, both SFARL$_{\text{BlurReg}}$ and SFARL$_{\text{NoiseReg}}$ are notably inferior to SFARL specified for deraning.
As shown in Fig. \ref{fig:filters_transfer}, most rain streaks can still be removed by SFARL$_{\text{BlurReg}}$ and SFARL$_{\text{NoiseReg}}$, but some fine-scale details may be blurry or smoothed out.
From the generative learning perspective, the regularization filters are used to model clean images, and can be transferred freely across tasks.
Nonetheless, due to the effect of discriminative learning, the regularization filters of SFARL are also tailored to the specific degradation type.
To sum up, regularization filters exhibit moderate generalization ability across different tasks, especially the two degradation types (e.g., deraining and denoising) are more similar.

Then, we transfer fidelity filters from deconvolution and denoising to deraining, denoted as SFARL$_{\text{BlurFid}}$ and SFARL$_{\text{NoiseFid}}$, respectively.
As shown in Table \ref{table:filters transfer} and Fig. \ref{fig:filters_transfer},  SFARL$_{\text{BlurFid}}$ and SFARL$_{\text{NoiseFid}}$ fail in removing rain streaks quantitatively and qualitatively.
Due to the correlation between fidelity filters and kernel estimation error, the result by SFARL$_{\text{BlurFid}}$ suffers from ringing effects.
The fidelity filters in SFARL$_{\text{NoiseFid}}$ are learned to model noises, and perform poor in removing rain streaks from rainy image.
Thus, fidelity filters are highly task-dependent, and cannot be transferred across tasks.

\begin{table}[!htb] \footnotesize
	\setlength{\abovecaptionskip}{1pt}
	\setlength{\belowcaptionskip}{0pt}
	\setlength{\tabcolsep}{2pt}
	\centering
	\caption{Quantitative results on rainy Rain12 dataset \cite{li2016rain} by transferring filters from SFARL models for deconvolution and denoising.} \label{table:filters transfer}
	\begin{tabular}{ccccccc}
		\hline
		
		\hline
		Method & SFARL & SFARL$_{\text{BlurReg}}$ & SFARL$_{\text{NoiseReg}}$ & SFARL$_{\text{BlurFid}}$ & SFARL$_{\text{NoiseFid}}$ \\
		\hline
		\hline
		PSNR &35.97 & 29.66 & 32.02 & 27.03 & 27.79 \\
		SSIM &0.9581& 0.8344& 0.9074& 0.6769 & 0.7508\\
		\hline
		
		\hline
	\end{tabular}
\end{table}
\begin{figure*}[!htb]\footnotesize
	\setlength{\abovecaptionskip}{1pt}
	\setlength{\belowcaptionskip}{0pt}
	\centering
	\setlength{\tabcolsep}{1pt}
	\begin{tabular}{ccccccc}
		\includegraphics[width=.24\textwidth]{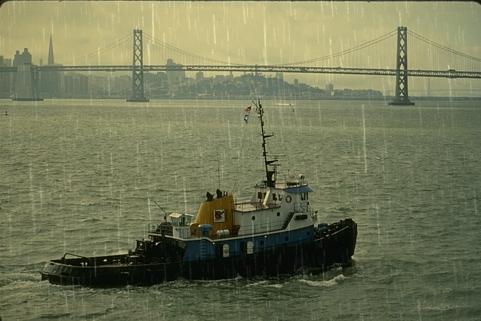}  &
		\includegraphics[width=.24\textwidth]{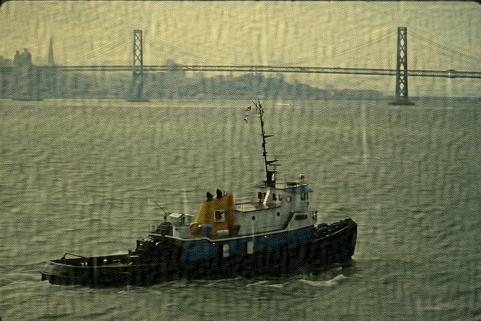}  &
		\includegraphics[width=.24\textwidth]{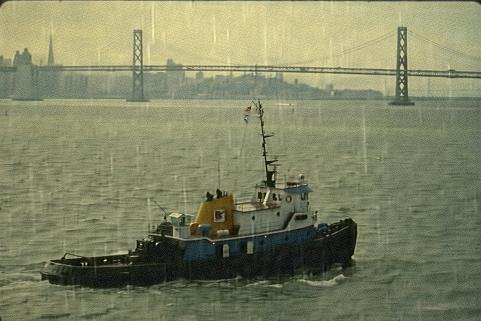}  \\
		Rainy image & SFARL$_{\text{BlurFid}}$ & SFARL$_{\text{NoiseFid}}$\\
		\includegraphics[width=.24\textwidth]{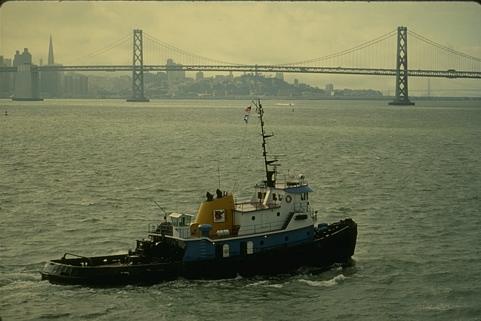}  &
		\includegraphics[width=.24\textwidth]{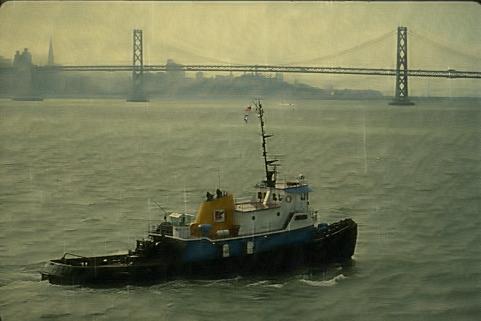}  &
		\includegraphics[width=.24\textwidth]{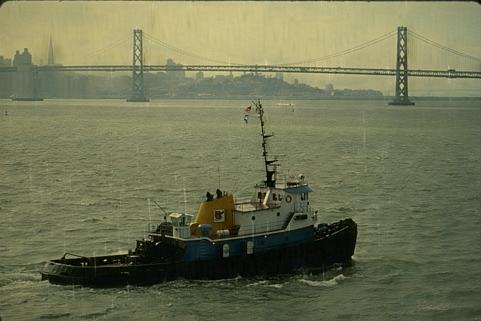}  \\
		Ground-truth & SFARL$_{\text{BlurReg}}$ & SFARL$_{\text{NoiseReg}}$ \\
	\end{tabular}
	\caption{Visual quality of deraining results by transferring filters from SFARL models for deconvolution and denoising.
	 }
	\label{fig:filters_transfer}
\end{figure*}

\vspace{-.1in}
\section{Conclusion}
\label{sec:conclusion}

In this paper, we propose an algorithm to effectively
handle image restoration with partially known or inaccurate degradation.
%
%
We present a flexible model to parameterize the fidelity term for characterizing spatial dependency and complex residual distribution of the residual image.
The simultaneous fidelity and regularization learning model is
developed by incorporating with the parameterized regularization term.
With a set of degraded and ground-truth image pairs, task-specific and stage-wise model parameters of SFARL can then be learned in a task driven manner.
Experimental results on two image restoration tasks, i.e., image deconvolution and rain streak removal, show that
the SFARL model performs favorably against
the state-of-the-art methods in terms of quantitative metrics and visual quality.
Experiments on Gaussian denoising show that the SFARL method is effective in improving visual perception metrics and visual quality of the denoising results.
%
%
Our future work includes extending the SFARL model to other restoration tasks, and
developing training methods within the unsupervised learning framework.

\section*{Acknowledgments}
This work is supported in part by National Natural Scientific Foundation of China (NSFC) under grant (61671182 and 61801326), and Hong Kong RGC GRF grant (PolyU 152124/15E), and US National Science Foundation CAREER Grant No.1149783.


%

\ifCLASSOPTIONcaptionsoff
  \newpage
\fi



%

%
%

\bibliographystyle{IEEEtran}
\bibliography{reference}
\vspace{-15mm}
\begin{IEEEbiography}[{\includegraphics[width=1in,height=1.25in,clip,keepaspectratio]{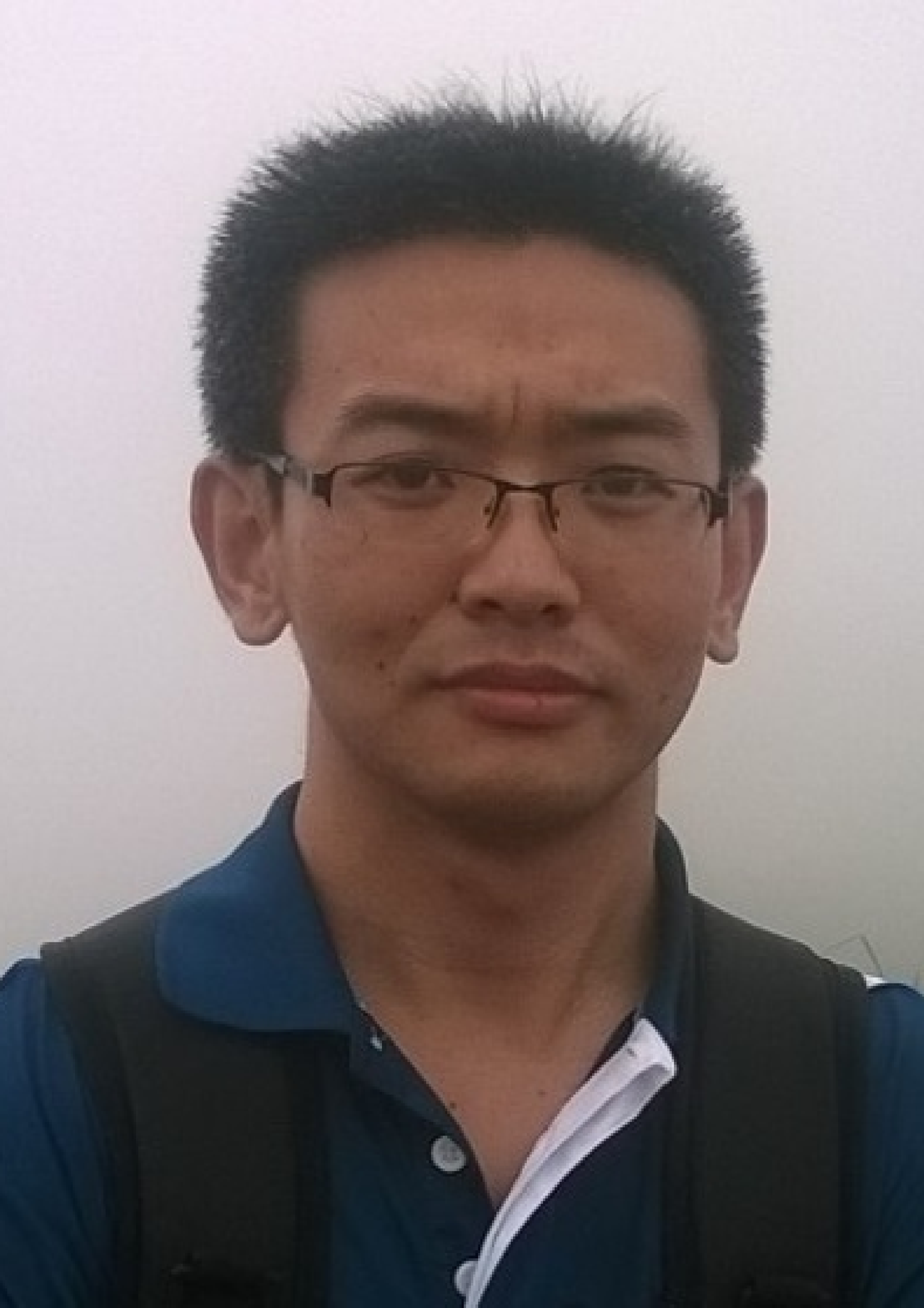}}]{Dongwei Ren}
received the Ph.D. degrees in computer application technology from Harbin Institute of Technology and The Hong Kong Polytechnic University in 2017 and 2018, respectively.
He is currently with the College of Intelligence and Computing, Tianjin University, China.
His research interests include low level vision, deep learning and optimization methods.
\end{IEEEbiography}
\vspace{-15mm}
\begin{IEEEbiography}[{\includegraphics[width=1in,height=1.25in,clip,keepaspectratio]{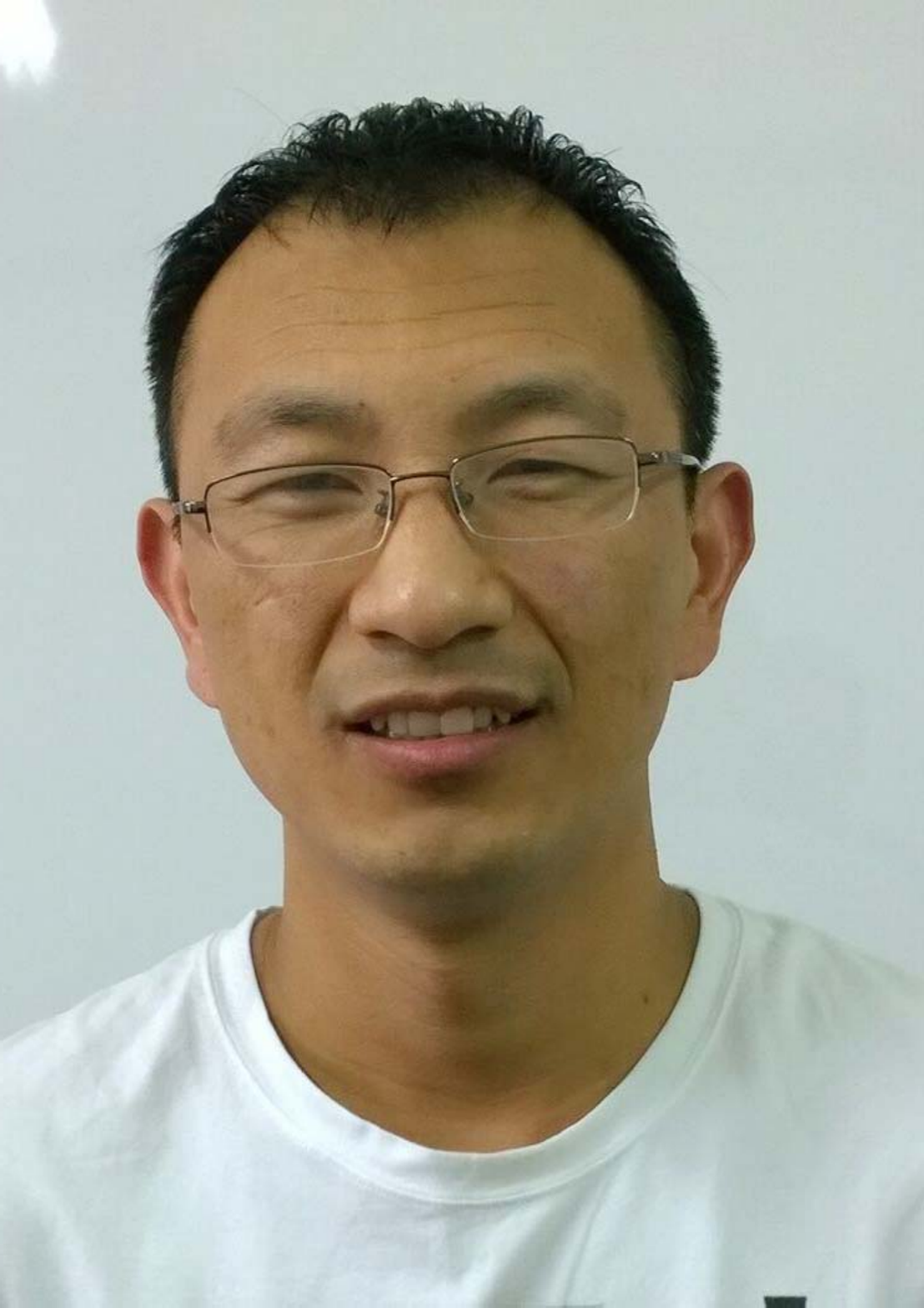}}]{Wangmeng Zuo} (M'09, SM'15)
received the Ph.D. degree in computer application technology from the Harbin Institute of Technology, Harbin, China, in 2007.
He is currently a Professor in the School of Computer Science and Technology, Harbin Institute of Technology. His current research interests include image enhancement and restoration, object detection, visual tracking, and image classification.
He has published over 70 papers in top-tier academic journals and conferences.
He has served as a Tutorial Organizer in ECCV 2016, an Associate Editor of the \emph{IET Biometrics} and \emph{Journal of Electronic Imaging}, and the Guest Editor of \emph{Neurocomputing}, \emph{Pattern Recognition}, \emph{IEEE Transactions on Circuits and Systems for Video Technology}, and \emph{IEEE Transactions on Neural Networks and Learning Systems}.
\end{IEEEbiography}
\vspace{-15mm}
\begin{IEEEbiography}[{\includegraphics[width=1in,height=1.25in,clip,keepaspectratio]{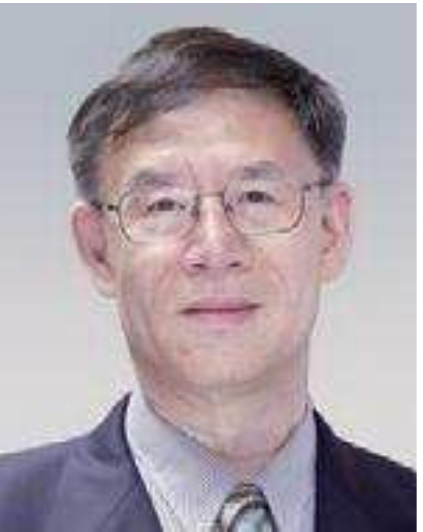}}]{David Zhang} graduated in Computer Science from Peking University. He received his MSc in 1982 and his PhD in 1985 in both Computer Science from the Harbin Institute of Technology (HIT), respectively. From 1986 to 1988 he was a Postdoctoral Fellow at Tsinghua University and then an Associate Professor at the Academia Sinica, Beijing. In 1994 he received his second PhD in Electrical and Computer Engineering from the University of Waterloo, Canada. He has been a Chair Professor at the Hong Kong Polytechnic University where he is the Founding Director of Biometrics Research Centre (UGC/CRC) supported by the Hong Kong SAR Government since 2005. Currently he is Presidential Chair Professor in Chinese University of Hong Kong (Shenzhen). He also serves as Visiting Chair Professor in Tsinghua University and HIT, and Adjunct Professor in Shanghai Jiao Tong University, Peking University and the University of Waterloo. He is both Founder and Editor-in-Chief, International Journal of Image \& Graphics (IJIG) and Springer International Series on Biometrics (KISB); Organizer, the first International Conference on Biometrics Authentication (ICBA); and Associate Editor of more than ten international journals including IEEE Transactions and so on. Over past 30 years, he has been working on pattern recognition, image processing and biometrics, where many research results have been awarded and some created directions, including palmprint recognition, computerized TCM and facial beauty analysis, are famous in the world. So far, he has published over 20 monographs, 450 international journal papers and 40 patents from USA/Japan/HK/China. He has been continuously listed as a Highly Cited Researchers in Engineering by Clarivate Analytics in 2014, 2015, 2016, 2017 and 2018, respectively. Professor Zhang is a Croucher Senior Research Fellow, Distinguished Speaker of the IEEE Computer Society, and a Fellow of both IEEE and IAPR.
\end{IEEEbiography}
\vspace{-16mm}
\begin{IEEEbiography}[{\includegraphics[width=1in,height=1.25in,clip,keepaspectratio]{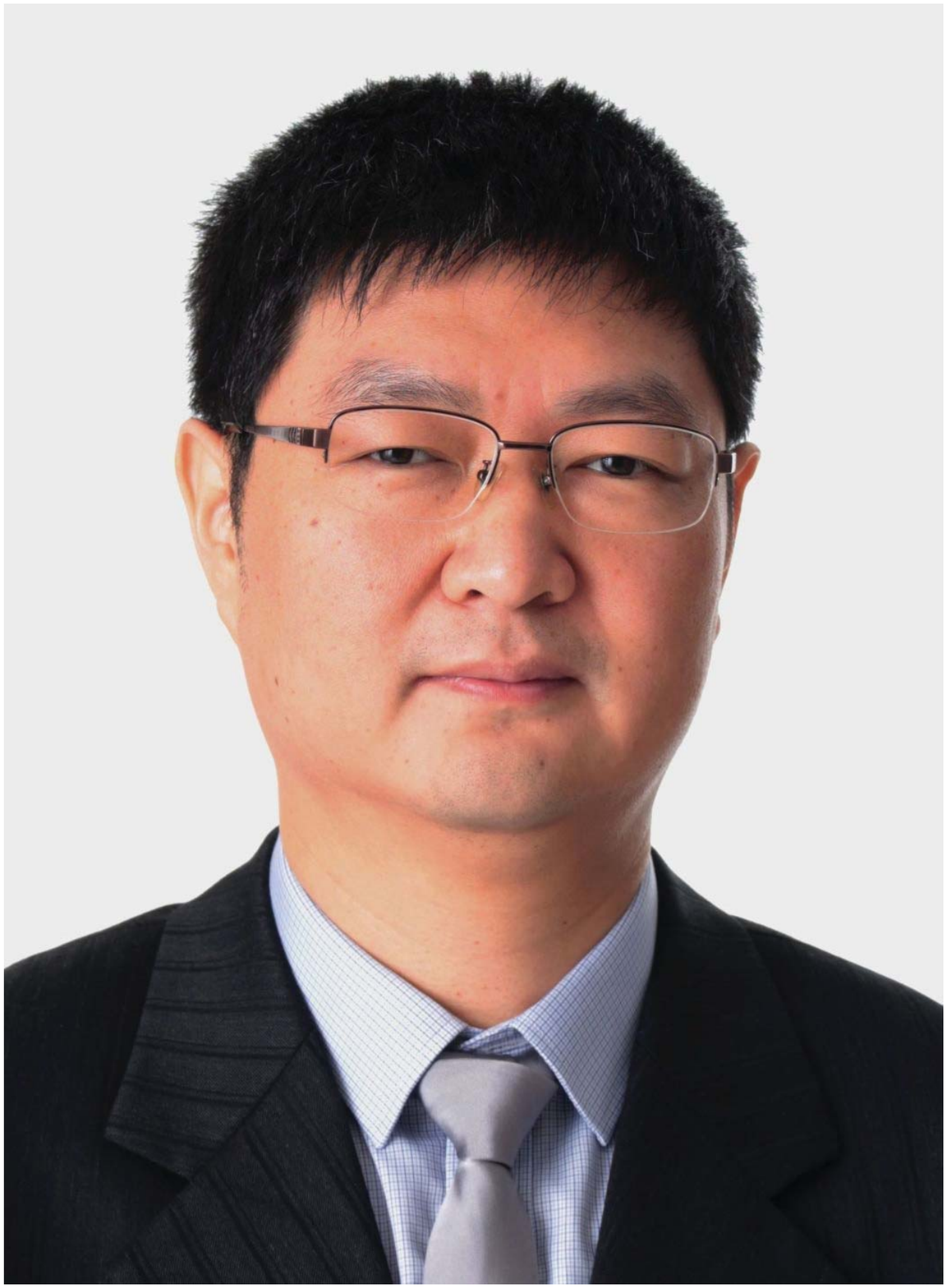}}]{Lei Zhang}
(M'04, SM'14, F'18) received his B.Sc. degree in 1995 from Shenyang Institute of Aeronautical Engineering, Shenyang, P.R. China, and M.Sc. and Ph.D degrees in Control Theory and Engineering from Northwestern Polytechnical University, Xi'an, P.R. China, in 1998 and 2001, respectively. From 2001 to 2002, he was a research associate in the Department of Computing, The Hong Kong Polytechnic University. From January 2003 to January 2006 he worked as a Postdoctoral Fellow in the Department of Electrical and Computer Engineering, McMaster University, Canada. In 2006, he joined the Department of Computing, The Hong Kong Polytechnic University, as an Assistant Professor. Since July 2017, he has been a Chair Professor in the same department. His research interests include Computer Vision, Image and Video Analysis, Pattern Recognition, and Biometrics, etc. Prof. Zhang has published more than 200 papers in those areas. As of 2019, his publications have been cited more than 40,000 times in literature. Prof. Zhang is a Senior Associate Editor of IEEE Trans. on Image Processing, and an Associate Editor of SIAM Journal of Imaging Sciences and Image and Vision Computing, etc. He is a "Clarivate Analytics Highly Cited Researcher" from 2015 to 2018. More information can be found in his homepage \url{http://www4.comp.polyu.edu.hk/~cslzhang/}.
\end{IEEEbiography}
\vspace{-15mm}
\begin{IEEEbiography}[{\includegraphics[width=1in,height=1.25in,clip,keepaspectratio]{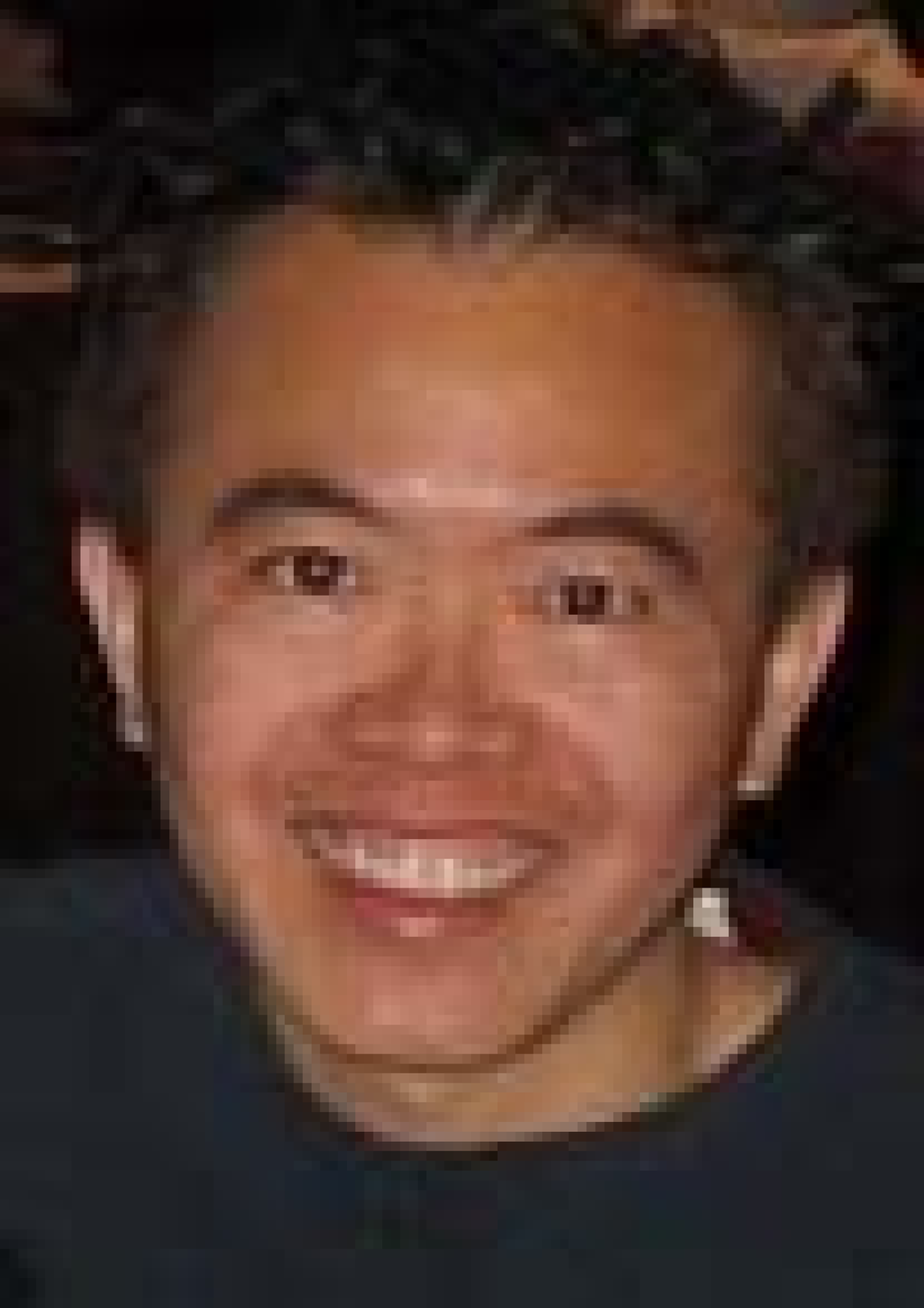}}]{Ming-Hsuan Yang}
is a Professor in Electrical Engineering and Computer Science at University of California, Merced. He received his PhD degree in computer science from the University of Illinois at Urbana-Champaign in 2000.
Yang served as an associate editor of the IEEE Transactions on Pattern Analysis and
Machine Intelligence from 2007 to 2011, and is an associate editor of the International Journal
of Computer Vision, Image and Vision Computing, and Journal of Artificial Intelligence Research. He received the Google Faculty Award in 2009 and the NSF CAREER Award in 2012. He is a senior member
of the IEEE and the ACM.
\end{IEEEbiography}
\vspace{-1cm}

\end{document}